\pdfoutput=1

\documentclass[11pt]{article}

\usepackage[]{acl}

\usepackage{times}
\usepackage{latexsym}

\usepackage[T1]{fontenc}

\usepackage[utf8]{inputenc}

\usepackage{microtype}

\usepackage{enumitem}
\usepackage{booktabs}
\usepackage{multirow}
\usepackage{lipsum} 
\usepackage{url}
\usepackage{graphicx}
\usepackage{caption}
\usepackage{subcaption}
\usepackage{svg}
\usepackage{layouts}

\def\easy{\emph{w/ cues }}
\def\hard{\emph{w/o cues }}

\def\overlap{\emph{lexical overlap}}
\def\subs{\emph{subsequence}}
\def\const{\emph{constituent}}

\newcommand\copa[3]{
\begin{description}[noitemsep]
    \item \textbf{Premise}: #1
    \item \textbf{a)} #2
    \item \textbf{b)} #3
\end{description}
}

\newcommand\copachoices[2]{
\begin{description}[noitemsep]
    \item \textbf{a)} #1
    \item \textbf{b)} #2
\end{description}
}

\newcommand\nli[3]{
\begin{description}[noitemsep]
    \item \textbf{Premise}: #1
    \item \textbf{Hypothesis}: #2
    \item \textbf{label}: #3
\end{description}
}

\title{Are Prompt-based Models Clueless?}

\author{%
    Pride Kavumba$^{1,2}$ \\ %
  \And
    Ryo Takahashi$^{3}$ \\
    $^1$Tohoku University %
    \quad $^2$RIKEN AIP %
    \quad $^3$LegalForce Research \\
    \texttt{kavumba.pride.q2@dc.tohoku.ac.jp} \\
    \texttt{\{ryo.takahashi, yusuke.oda\}@legalforce.co.jp} \\ %
  \And
    Yusuke Oda$^{3,1}$ \\ %
 }

\begin{document}
\maketitle
\begin{abstract}
Finetuning large pre-trained language models with a task-specific head has advanced the state-of-the-art on many natural language understanding benchmarks. However, models with a task-specific head require a lot of training data, making them susceptible to learning and exploiting dataset-specific superficial cues that do not generalize to other datasets.
Prompting has reduced the data requirement by reusing the language model head and formatting the task input to match the pre-training objective.
Therefore, it is expected that few-shot prompt-based models do not exploit superficial cues.
This paper presents an empirical examination of whether few-shot prompt-based models also exploit superficial cues.
Analyzing few-shot prompt-based models on MNLI, SNLI, HANS, and COPA has revealed that prompt-based models also exploit superficial cues.
While the models perform well on instances with superficial cues, they often underperform or only marginally outperform random accuracy on instances without superficial cues.
\end{abstract}

\section{Introduction}
\label{sec:introduction}

Finetuning large pre-trained language models with a task-specific head has achieved remarkable performance on many natural language benchmarks~\cite{wang-etal-2018-glue, wang2019superglue}.
However, the task-specific head introduces a lot of random task-specific parameters that require enormous finetuning data to attain optimal performance. The exposure to enormous data increases the potential for models to learn and exploit dataset-specific superficial cues that do not generalize to other datasets without superficial cues~\cite{gururangan-etal-2018-annotation, poliak-etal-2018-hypothesis, sugawara-etal-2018-makes, niven2019probing, schuster-etal-2019-towards, kavumba-etal-2019-choosing}.
For example, \citet{niven2019probing} found that task-specific head models exploit the presence of ``not'' in the input of argument reasoning comprehension dataset~\cite{habernal-etal-2018-semeval-arct} to achieve state-of-the-art accuracy, but drop to random accuracy when the superficial cue is neutralized.
On the other hand, prompting reuses the pre-training language model head, introducing no random task-specific parameters.
Thus, prompt-based models can achieve remarkable performance with only a few training examples~\cite{gpt3_NEURIPS2020_1457c0d6, schick-schutze-2021-exploiting, schick-schutze-2021-just, gao-etal-2021-making-lm-bff, le-scao-rush-2021-many-data-points}.
Hence, few-shot prompting lowers the potential for models to learn and exploit dataset-specific superficial cues.

\begin{figure}[t]
\resizebox{\linewidth}{!}{%
\includegraphics[width=8cm]{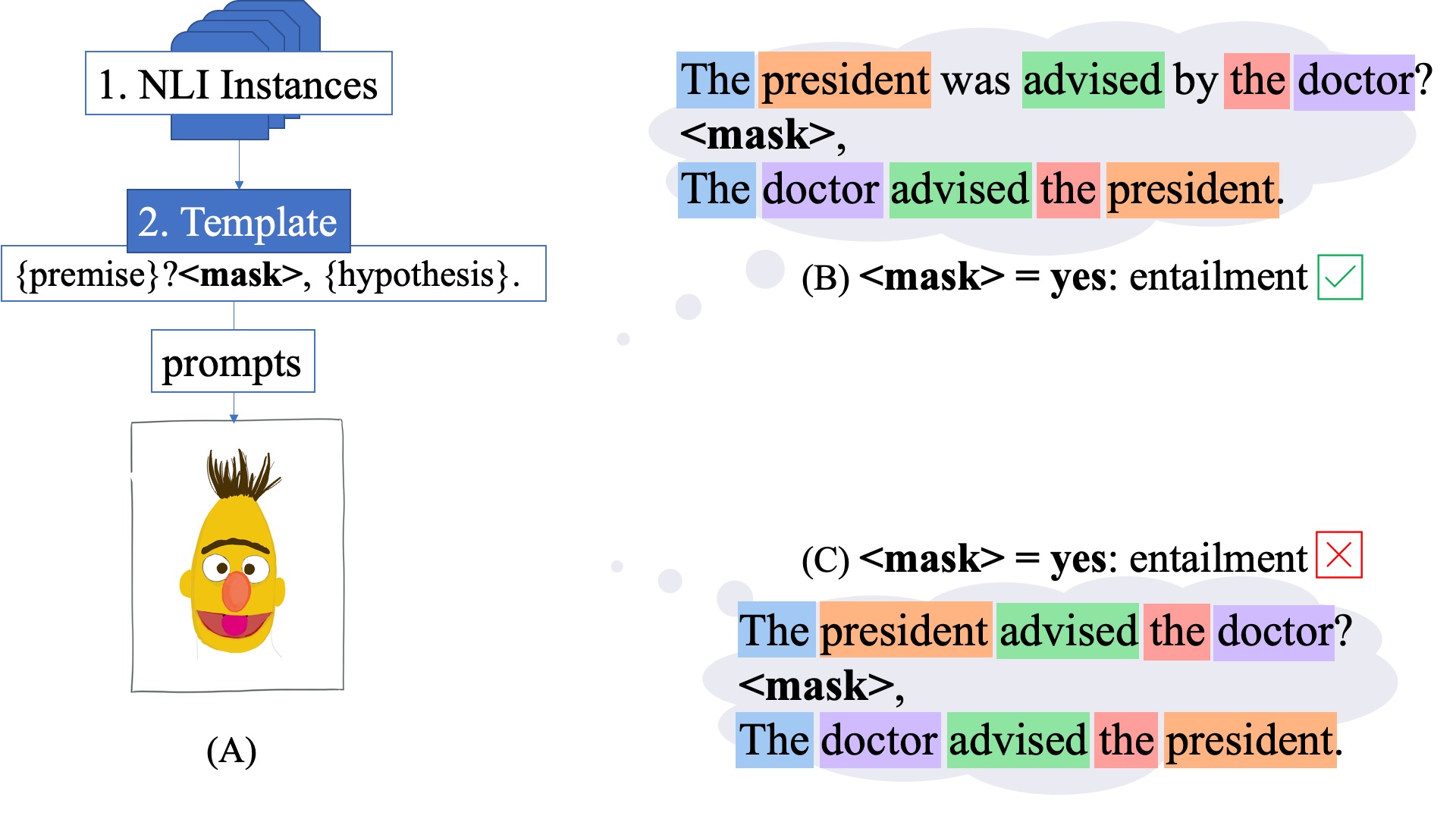}}
\caption{(A) shows a prompt-based model receiving natural language inference (NLI) prompts generated through a template.
(B) Our analysis reveals that prompt-based models exploit superficial cues (highlighted lexical overlap) that are predictive of entailment relation between premise and hypothesis.
(C) The model fails to generalize to instances were the superficial cues no longer predict entailment relation.}
\label{fig:concept}
\end{figure}

This work empirically investigates whether few-shot prompt-based models exploit superficial cues. Specifically, we ask: \emph{Do few-shot prompt-based models exploit superficial cues?}
To answer this question, we examine prompted-based models on two fundamental tasks of natural language understanding: natural language inference (NLI) and commonsense reasoning;
comprehending natural language inference and commonsense are essential to make progress in natural language understanding~\cite{bowman-etal-2015-large-snli, williams-etal-2018-broad, roemmele2011choice}.
We analyze the performance of prompt-based models trained on the Stanford Natural Language Inference dataset~\cite[SNLI]{bowman-etal-2015-large-snli}, the Multi-Genre Natural Language Inference data~\cite[MNLI]{williams-etal-2018-broad}, and the Choice of Plausible Alternatives dataset~\cite[COPA]{roemmele2011choice} on instances with and without superficial cues.

To facilitate the analysis, we define two types of superficial cues that abstract away from the underlying tasks: \emph{context} and \emph{contextless} superficial cues, where the definition of the \emph{context} depends on the task.
For example, in natural language inference tasks, we define the premise as the \emph{context}, while in multiple-choice tasks, we define the question as the \emph{context}.
Context superficial cues such as lexical overlap (Figure~\ref{fig:concept}) coexist in the context (premise) and the hypothesis. In contrast, contextless superficial cues exist only in the hypothesis (in NLI) or in answer choices (in multiple-choice tasks).
A dataset can contain either one or both types of superficial cues.
Therefore, both types must be investigated to sufficiently answer whether a model can exploit superficial cues.

As a prerequisite, we reanalyze superficial cues in MNLI, SNLI, and COPA datasets to created evaluation sets that have and do not have superficial cues.
We find that these datasets contain more superficial cues than previously known.
Specifically, we find that 90.1\% of MNLI matched instances contain \emph{contextless superficial cues} in the hypothesis, while 71.9\% SNLI contains \emph{contextless superficial cues}.
Additionally, we find that COPA contains not only contextless superficial cues~\cite{kavumba-etal-2019-choosing}, but 78.0\% of the instances also contain \emph{context superficial cues}.

Finally, we examine whether few-shot prompt-based models also rely on superficial cues to achieve remarkable performance on MNLI, SNLI, COPA, and the HANS dataset~\cite{mccoy2019right}.
COPA experiments reveal that prompt-based models do not rely on contextless superficial cues for typical few-shot training sizes.
However, the other empirical results show that prompt-based models heavily rely on superficial cues---failing to generalize to data without superficial cues (Figure~\ref{fig:concept}).

In summary, our contributions are:
\begin{enumerate}[noitemsep]
    \item We propose to divide superficial cues into \emph{context} and \emph{contextless} superficial cues, which abstracts away from the underlying tasks (\S~\ref{sec:superficial-cues-datasets}).
    \item We established that the datasets of MNLI~(
    \S~\ref{sec:contextless-superficial-cues}), SNLI~(\S ~\ref{sec:contextless-superficial-cues}) and COPA~(\S ~\ref{sec:context-superficial-cues}) contain more superficial cues than previously known. We release analyzed datasets at \url{https://github.com/legalforce-research/prompt-models-clueless}.
    \item We present the first investigation of the exploitation of superficial cues by prompt-based models, finding that prompt-based models also exploit superficial cues~(\S~\ref{sec:experiments}). 
\end{enumerate}

\section{Background}
\label{sec:background}

We will begin with a review of the necessary concepts required to understand the rest of the paper.

\subsection{Prompting}
\label{sec:prompting}

Prompt-based finetuning has been demonstrated to be effective in few-shot setup~\cite{gpt3_NEURIPS2020_1457c0d6, schick-schutze-2021-exploiting, schick-schutze-2021-just, gao-etal-2021-making-lm-bff, le-scao-rush-2021-many-data-points}.
By reusing the pretraining language model head, prompting introduces no or only a few randomly initialized parameters.
Prompting reformulates any task to match the pretraining objective.
For example, consider the task of classifying the sentiment polarity of movie reviews using a masked language model such as BERT~\cite{devlin-etal-2019-bert}.
A review such as ``\emph{I liked the movie}'' is converted to ``\emph{I liked the movie. It was [MASK]}''.
The model, then, fills [MASK] with words such as \emph{\{good, nice, bad, terrible\}}, which are mapped to the task labels---positive or negative---through a verbalizer~\cite{schick-schutze-2021-exploiting, schick-schutze-2021-just}.
In contrast, a task-specific classification head model directly predicts positive or negative sentiment.
For an in-depth review, we will refer the interested reader to a survey by \citet{liu2021pretrain-prompt-survey}.

\subsection{Superficial Cues}
\label{sec:superficial-cues}

Superficial cues can be described as linguistic or non-linguistic characteristics of instances that have nothing to do with the task itself but are tied to a specific task label.
These characteristics include lexical overlap~\cite{mccoy2019right}, distinct words frequently appearing in the correct choices~\cite{niven2019probing, kavumba-etal-2019-choosing}, and distinctive style of the correct choices~\cite{trichelair-etal-2019-reasonable}.
As a concrete example, consider a sentiment classification dataset whose negative sentiment instances contain ``not''; for example, ``I did not like the movie''.
Here, ``not'' is a superficial cue because it is predictive of the correct label.

\subsection{Training Datasets}
\label{sec:datasets}

\paragraph{MNLI} The Multi-Genre Natural Language Inference~\cite[MNLI]{williams-etal-2018-broad} dataset is an important dataset of natural language inference which is also part of the SuperGLUE benchmark~\cite{wang2019superglue}.
Given a premise and a hypothesis, the task asks to pick one label from among three, \{contradiction,  neutral, entailment\}.
The test set of MNLI is divided into matched (in-domain instances) and mismatched (out-of-domain instances) subsets based on whether the domain of each test instance matches the training set domain.

\paragraph{SNLI} The Stanford Natural Language Inference~\cite[SNLI]{bowman-etal-2015-large-snli} is a popular natural language inference dataset with the same format as MNLI.

\paragraph{COPA} The Choice of Plausible Alternatives~\cite[COPA]{roemmele2011choice} dataset is a popular multiple-choice commonsense dataset, which is also a part of the SuperGLUE benchmark. Given a premise and a question, the task is to select the most plausible cause or effect from the set of two candidates.

\section{Superficial Cues in NLI and COPA}
\label{sec:superficial-cues-datasets}

We investigate prompted-based models on two fundamental tasks of natural language understanding: natural language inference (NLI) and commonsense reasoning.
As a prerequisite, we begin by creating test sets with and without superficial cues that we will subsequently use to investigate whether prompt-based models exploit superficial cues.
We analyze and split test sets of English language datasets into subsets with and without superficial cues in the following subsections.

To facilitate easy analysis, we divide superficial cues into two categories: \emph{context} superficial cues and \emph{contextless} superficial cues, where the definition of \emph{context} is task dependent.
For example, in natural language inference tasks such as COPA, the \emph{context} can be defined as the premise, while in multiple-choice tasks, the \emph{context} can be defined as the question.
Context superficial cues, such as lexical overlap found by \citet{mccoy2019right} can only be exploited when the context is available in the input.
On the other hand, contextless superficial cues, such as the occurrence of ``not'' in the correct answer choices found by \citet{niven2019probing}, are those that are exploitable even in the absence of the context required to perform a task.

\subsection{Context Superficial Cues}
\label{sec:context-superficial-cues}

\paragraph{Natural Language Inference (NLI):} 
NLI has a good dataset designed to test for contextless superficial cues. Specifically, the HANS dataset tests the models' ability to exploit
three types of context superficial cues in NLI: \overlap{}, \subs{}, and \const{}~\citet{mccoy2019right}.
Therefore, we evaluate prompt-based models on the HANS dataset instead of splitting tests of MNLI and SNLI into instances with and without superficial cues.

\paragraph{COPA}
Eyeballing all instances to find common patterns that identify the correct answer choice, but are unrelated to the task, can be challenging and error-prone.
To circumvent the need for manual examination, we propose to solve the task in a setup that encourages the model to solve the task using superficial cues. 
This setup is similar to providing only partial input~\cite{gururangan-etal-2018-annotation,poliak-etal-2018-hypothesis}.
Specifically, we randomly shuffle the words in the answer choices such that identifying the correct choice is mainly based on superficial cues in the question and the answer choice.
For example, given the original instance;

\copa
{The host cancelled the party. What was the CAUSE of this?}
{She worried she would catch the flu.}
{She was certain she had the flu. (correct)}

\noindent
The new answer choices for the new instance becomes:
\copachoices
{She would she catch the worried flu.}
{She had was she the certain flu. (correct)}

In this setting, we find that RoBERTa achieves an average accuracy of 78\%, indicating the existence of context superficial cues.
Following this result, we split the test set into a subset with superficial cues containing instances solved by the majority of models, and a subset without superficial cues, containing all the remaining instances.

\subsection{Contextless Superficial Cues}
\label{sec:contextless-superficial-cues}

\paragraph{Natural Language Inference}

\begin{table}[t]
\centering
\begin{tabular}{@{}lr@{}}
\toprule
Dataset & \multicolumn{1}{c}{accuracy}           \\ \midrule
Random & 33.3 \\ \cmidrule(l){1-2}
MNLI    & 90.1 \small{$\pm$0.1} \\
MNLI-mm & 90.0 \small{$\pm$0.2} \\ \cmidrule(l){1-2}
SNLI    & 71.9 \small{$\pm$0.1} \\ \bottomrule
\end{tabular}
\caption{Average accuracy on matched MNLI (MNLI) and mismatched MNLI (MNLI-mm), and SNLI for a head RoBERTa model trained on the hypothesis.
MNLI and MNLI-mm results are for a model trained on MNLI and SNLI results are for a model trained on SNLI.}
\label{tab:head-contextless-acc}
\end{table}

To investigate contextless superficial cues in NLI,
we train RoBERTa~\cite{RoBERTa2019} with a classification head on only the hypothesis of MNLI and SNLI.
This analysis is similar to the one done by \citet{gururangan-etal-2018-annotation} using fastText~\cite{fastText-joulin2017bag}.
If the model can not find contextless superficial cues in the hypothesis, it is expected to achieve random performance (33.3\%).
But, RoBERTa trained on MNLI achieve an average performance of 90.1\% and 90.0\% on matched and  mismatched instances, respectively~(Table~\ref{tab:head-contextless-acc}), which is worse than previously known (53.9\% matched and 52.3\% mismatched~\cite{gururangan-etal-2018-annotation}).
On the test set of SNLI, RoBERTa trained on SNLI achieves an average accuracy of 71.9\%~(Table~\ref{tab:head-contextless-acc}), which is 4.9 percentage points higher than previously known \citet{gururangan-etal-2018-annotation}.
Following this result, we split the testing sets of MNLI and SNLI such that each test set has two subsets: instances with contextless superficial cues, containing all instances that the majority of models solved correctly, and instances without contextless superficial cues contain all the remaining instances.

\begin{figure*}[t!]
    \centering
    \begin{subfigure}[t]{0.6\textwidth}
        \centering
        \includegraphics[width=\textwidth]{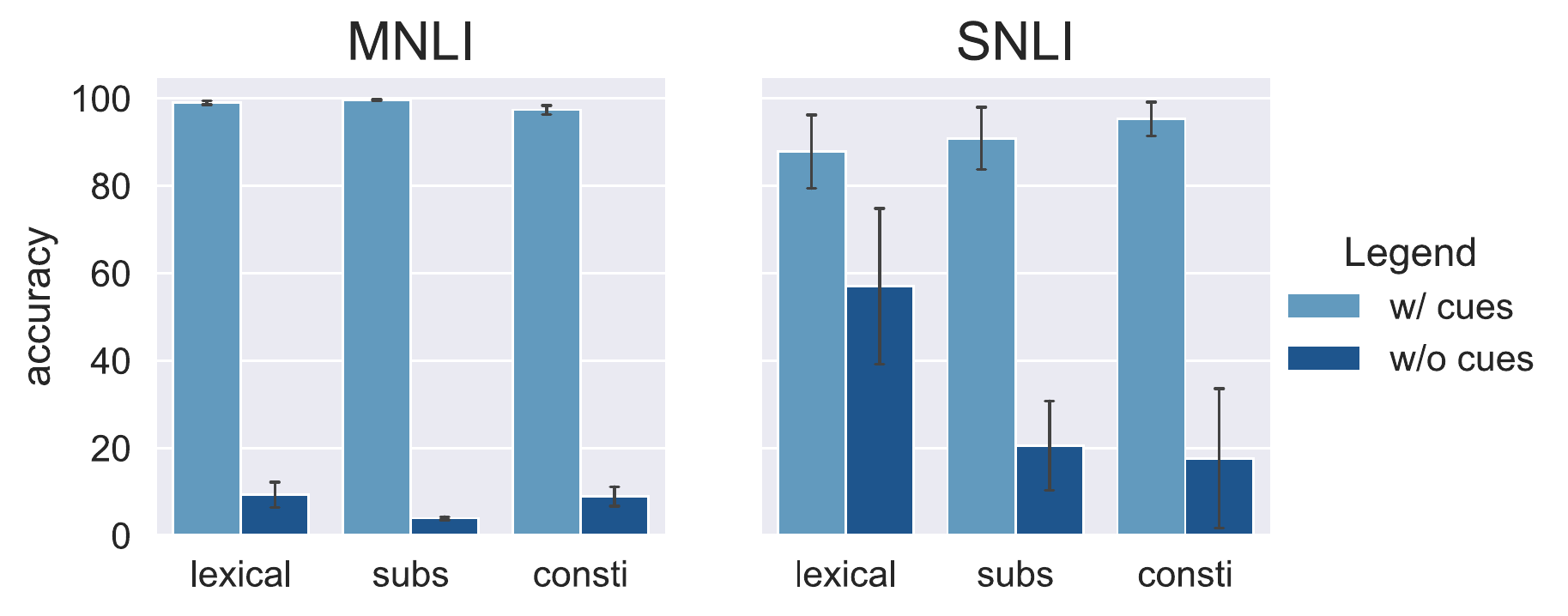}
        \caption{w/ \& w/o Context Cues}
        \label{fig:mnli-snli-hans-results}
    \end{subfigure}%
    ~ 
    \begin{subfigure}[t]{0.4\textwidth}
        \centering
        \includegraphics[width=\textwidth]{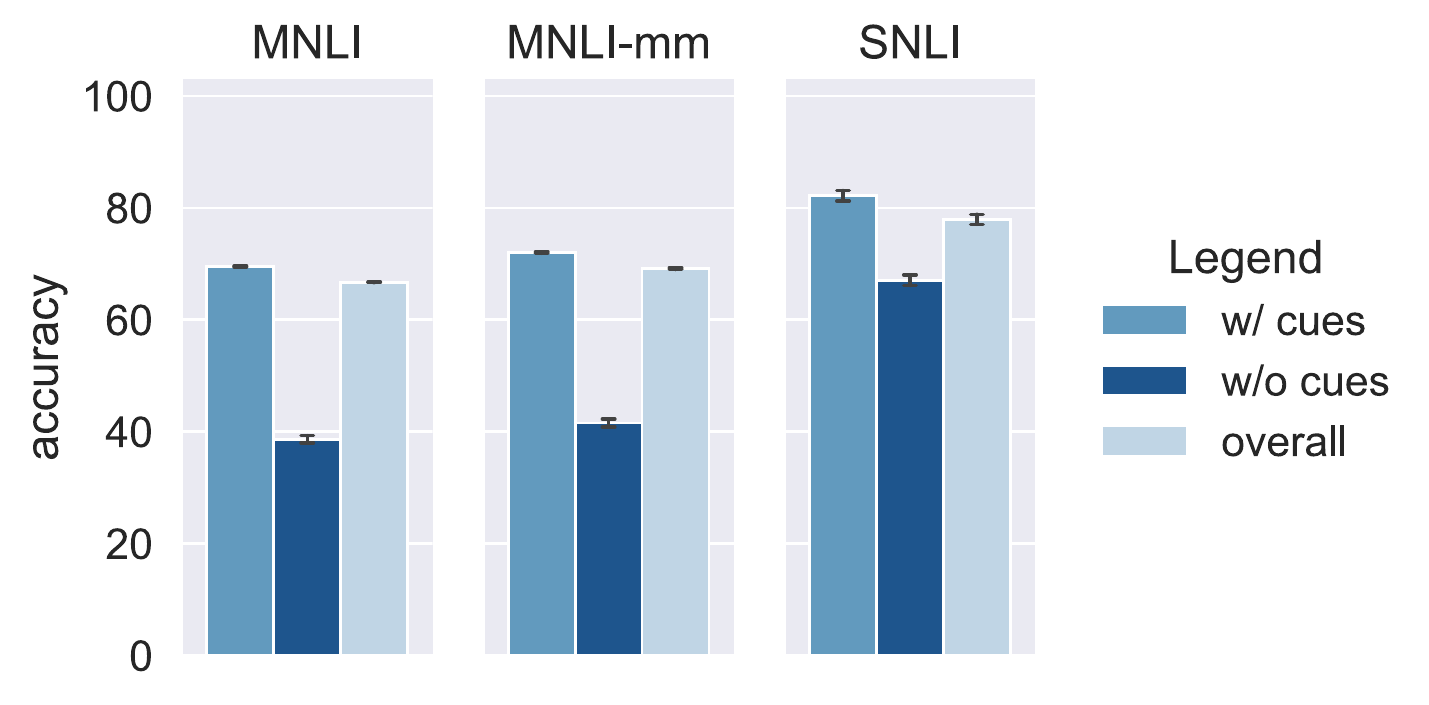}
        \caption{w/ \& w/o Contextless Cues}
        \label{fig:mnli-snli-contextless-results}
    \end{subfigure}
    \caption{Average accuracy on instances with superficial cues (w/ cues) and without superficial cues (w/o cues) for prompt-based RoBERTa models trained on MNLI and SNLI.
    (a) shows results on the HANS dataset on three kinds of context superficial cues: lexical overlap (lexical), subsequence (subs), and constituent (consti).
    (b) shows results on instances with and without contextless superficial cues of matched MNLI (MNLI) and mismatched MNLI (MNLI-mm) for a model trained on MNLI; and evaluations results on SNLI for a model trained on SNLI.}
\end{figure*}

\paragraph{COPA} The test set of COPA has already been split into two subsets that have instances with contextless superficial cues and instances that do not have contextless superficial cues~\citet{kavumba-etal-2019-choosing}.
The subsets were constructed based on the performance of RoBERTa trained on answers only.
Therefore we do not reanalyze COPA; instead, we will use the same publicly available subsets in our evaluation.

\section{Experimental Setup}
The goal of our experimental setup is to answer the following research question: Do prompt-based models exploit superficial cues?
We decompose this question into two sub-questions: 1) Do prompt-based models exploit context superficial cues? 2) Do prompt-based models exploit contextless superficial cues?

\paragraph{Training Details}
\label{sec:training-details}

For all our experiments, we use RoBERTa-large (355M parameters) because it is the widely used model in prompt-based finetuning~\cite{schick-schutze-2021-exploiting, gao-etal-2021-making-lm-bff, le-scao-rush-2021-many-data-points}.
We build on the source code by \citet{gao-etal-2021-making-lm-bff}~\footnote{\url{https://github.com/princeton-nlp/LM-BFF}} and \citet{le-scao-rush-2021-many-data-points}~\footnote{\url{https://github.com/TevenLeScao/pet}}, and we load the pre-trained weights from HuggingFace~\cite{Wolf2019HuggingFacesTS}.
We use the best-reported hyperparameters and templates~(Appendix~\ref{app:prompt-templates}) from~\citet{gao-etal-2021-making-lm-bff} on NLI.
All NLI models are trained with 16 instances per label.
We use the same partitions used by~\citet{gao-etal-2021-making-lm-bff}.
On COPA we use the best hyperparameters and templates~(Appendix~\ref{app:prompt-templates}) from~\citet{schick-schutze-2021-just,le-scao-rush-2021-many-data-points}.
We ran all experiments three times with different random seeds and report the average and standard deviation.

\paragraph{Evaluation}
The goal of our experimental evaluation is to answer the following question: Do prompt-based models exploit superficial cues? We answer this question by investigating whether the model exploits or relies on either context or contextless superficial cues. We train and evaluate our models on English datasets.

\paragraph{Context Superficial Cues}
To investigate whether models exploit context superficial cues, we train prompt-based models on MNLI, SNLI, and COPA.
We evaluate NLI models on the HANS dataset that tests the models' ability to exploit
three types of context superficial cues in NLI: \overlap{}, \subs{}, and \const{}~\citet{mccoy2019right}.
We report the average accuracy and standard deviation on the two subsets of the dataset: a subset where the superficial is informative (Entailment) and a subset where the superficial cues are uninformative (Non-entailment).
A model that does not rely on context superficial cues is expected to perform comparably on both subsets.

\paragraph{Contextless Superficial Cues}
To investigate whether prompt-based models exploit contextless superficial cues, we train a prompt-based model on MNLI, SNLI, and COPA; and evaluate them on the corresponding test set of each dataset. Each test set consists of two subsets obtained and described in sections~\ref{sec:superficial-cues-datasets}: a subset of instances with contextless superficial cues and a subset without contextless superficial cues.
A model that does not rely on contextless superficial cues is expected to perform comparably on both subsets.

\section{Results}
\label{sec:experiments}

\subsection{Exploiting Context Superficial Cues}
\label{sec:results-context-supericial-cues}

\paragraph{Natural Language Inference (NLI)}

Figure~\ref{fig:mnli-snli-hans-results} shows the results on the HANS dataset of the prompt-based model trained on MNLI (left) and SNLI (right), respectively.
The results show that prompt-based RoBERTa trained on MNLI performs considerably well on instances with superficial cues, an overall average of 98.7\%.
However, the model only achieves an overall average accuracy of 7.4\% on instances without context superficial cues, failing to reach random accuracy of 50\%.
This indicates that the prompt-based models trained on MNLI exploit superficial cues.

Similarly, figure~\ref{fig:mnli-snli-hans-results} (right) shows that while RoBERTa performs considerably well on instances with superficial cues (overall average 91.3\%), it fails to achieve the same performance on instances without superficial cues (overall average of 31.7\%).~\footnote{The high variance is similar to that reported by previous work studying head models~\cite{pmlr-v119-bras20a-aflite}.}
This result also leads to the same conclusion that the model exploits context superficial cues.

\begin{table}[t]
\centering
\begin{tabular}{@{}lllll@{}}
\toprule
\# Examples & \easy                  & \hard                 & Overall                 &                 \\ \midrule
8           & 83.0{\small $\pm$0.5} & 55.7{\small $\pm$1.4} & \multicolumn{2}{l}{78.1{\small $\pm$0.3}} \\
16          & 81.6{\small $\pm$1.9} & 57.9{\small $\pm$1.0} & \multicolumn{2}{l}{77.3{\small $\pm$1.4}} \\
32          & 82.4{\small $\pm$2.0} & 53.5{\small $\pm$0.5} & \multicolumn{2}{l}{77.1{\small $\pm$1.5}} \\
64          & 84.4{\small $\pm$1.4} & 53.8{\small $\pm$4.1} & \multicolumn{2}{l}{78.9{\small $\pm$1.2}} \\
96          & 87.0{\small $\pm$1.7} & 57.9{\small $\pm$4.1} & \multicolumn{2}{l}{81.7{\small $\pm$0.8}} \\
100         & 87.9{\small $\pm$1.7} & 54.2{\small $\pm$1.9} & \multicolumn{2}{l}{81.7{\small $\pm$1.2}} \\ \bottomrule
\end{tabular}
\caption{Average accuracy of few-shot prompt-based RoBERTa models on COPA instances with context superficial cues between the context and the answer choices; and instances without exploitable context superficial cues.
Column ``\# Examples'' shows the number of examples used for finetuning.}
\label{tab:copa-context-superficial-cues}
\end{table}

\paragraph{COPA}
Table~\ref{tab:copa-context-superficial-cues} shows the results of the prompt-based RoBERTa trained on COPA and evaluated on the two subsets of COPA: with and without superficial cues. The results show RoBERTa performs well on instances with superficial cues but barely exceeds random accuracy (50\%) on instances without superficial cues.
This, too, indicates that the model exploit contextless superficial cues.

\subsection{Exploiting Contextless Superficial Cues}

\paragraph{NLI} Figure~\ref{fig:mnli-snli-contextless-results} shows the results of prompt-based RoBERTa train on MNLI and SNLI and evaluated on the corresponding test set.
The results show that the prompt-based model trained on MNLI performs considerably better on instances with superficial cues on both matched (69.5\%) and mismatched (72.0\%) instances.
On instances without superficial cues we observe a gap of 30.9\% and 30.5\% on matched and mismatched instances, respectively.
The high difference in performance indicates that the models exploit contextless superficial cues.
For the model trained on SNLI and evaluated on SNLI subsets, we observe a gap of 15.1\% between performance on instances with and without superficial cues (82.2\% vs 67.1\%). 
This also indicates that the model does exploit contextless superficial cues.

\paragraph{COPA}
\begin{table}[t]
\centering
\begin{tabular}{@{}llll@{}}
\toprule
\# Examples & \easy                & \hard                 & Overall               \\ \midrule
8           & 79.1{\small $\pm$0.9} & 77.4{\small $\pm$0.0}   & 78.1{\small $\pm$0.3} \\
16          & 77.0{\small $\pm$1.3}   & 77.4{\small $\pm$1.6} & 77.3{\small $\pm$1.4} \\
32          & 77.7{\small $\pm$3.1} & 76.8{\small $\pm$0.9} & 77.1{\small $\pm$1.5} \\
64
& 80.2{\small $\pm$0.5} & 78.1{\small $\pm$1.6} & 78.9{\small $\pm$1.2} \\
96          & 83.7{\small $\pm$1.1} & 80.4{\small $\pm$0.7} & 81.7{\small $\pm$0.8} \\
100         & 84.6{\small $\pm$1.1} & 80.0{\small $\pm$1.6}   & 81.7{\small $\pm$1.2} \\ \bottomrule
\end{tabular}
\caption{Average accuracy of few-shot prompt-based RoBERTa models on COPA instances with contextless superficial cues (w/ cues) in the answer choices and instances without contextless superficial cues (w/o cues).
Column ``\# Examples'' shows the number of examples used for finetuning.}
\label{tab:copa-contextless-superficial-cues}
\end{table}

Table~\ref{tab:copa-contextless-superficial-cues} shows of the prompt-based RoBERTa trained on COPA and evaluated on the COPA subsets with and without superficial cues.
The results show that the prompt-based model does not exploit superficial cues at a small enough training set (less or equal to 32 instances).
However, increasing the size further increases the gap in performance between instances with and without contextless superficial cues.
It is encouraging to note that the model does not exploit contextless superficial cues at sizes commonly used in few-shot settings.

\section{Discussion}
\label{sec:discussion}

\subsection{Predictions Errors}
\label{sec:prediction-analysis}

The results on natural language inference instances without superficial cues are worse than random performance.
One wonders whether it is because the instances are hard.
We look at some instances that the model fails to solve correctly.
We show some of the instances in Table~\ref{tab:mnli-wrong-predictions}.
The instances are simple enough for anyone that understands English. 
One question that immediately arises is; are prompt-based models sensitive to the meaning of the question?

\begin{table*}[t]
\centering
\begin{tabular}{@{}llll@{}}
\toprule
Premise                              & Hypothesis                        & Label & Prediction\\ \midrule
The president advised the doctor     & The doctor advised the president  & N & E \\
The student saw the managers         & The managers saw the student      & N & E \\
The presidents encouraged the banker & The bank encouraged the president & N & E \\
The actors avoided the bankers       & The bankers avoided the bankers   & N & E \\
The managers saw the secretaries     & The secretary saw the managers    & N & E \\
The lawyers helped the judge         & The judge helped the lawyers      & N & E \\
The banker thanked the tourist       & The tourist thanked the banker    & N & E \\ \bottomrule
\end{tabular}
\caption{Some examples of instances without superficial cues from the HANS dataset that a prompt-based model trained on MNLI wrongly classify as entailment (E) because of superficial cues: high lexical overlap.}
\label{tab:mnli-wrong-predictions}
\end{table*}

\subsection{Attention Visualization}
\label{sec:attention-analysis}

To investigate whether prompt-based models are sensitive to meaning, we compare the attention weight across all twenty-four layers of RoBERTa-large for closely related instances that differ only in meaning and hence the labels.
While there have been many questions that have been raised over the reliability of singly using attention weights for explanation~\cite{wiegreffe-pinter-2019-attention-not-explanation, vig-belinkov-2019-analyzing-attention}, here we use attention weights coupled with other results to gain more insights into the models' inner working.
We are interested in knowing whether there is a huge change in attention weights responding to the change in meaning.
For example, we take an instance with superficial cues, which lead to the correct prediction of the Entailment label:

\nli
{The president was advised by the doctor.}
{The doctor advised the president.}
{Entailment}

And an instance without superficial cues:
\nli
{The president advised the doctor.}
{The doctor advised the president.}
{Non-Entailment}

While the instances are completely different in meaning, the model predicts Entailment in both cases because of the superficial cue of high overlap.
When we compare the attention maps for all the layers, we notice that there is barely any change in response to the change in the meaning of the sentences.
Because of space limitation, we show attention maps only for the first two layers and the last layer (Figure~\ref{fig:attention-maps-hans}).
The attention maps for all the 24 layers are shown in Appendix~\ref{app:attention}.
The visualizations highlight the inability of the model to respond to the change in meaning.
We investigate this further in the next section.

\begin{figure*}[t]
\resizebox{\linewidth}{!}{%
\includegraphics[width=\textwidth]{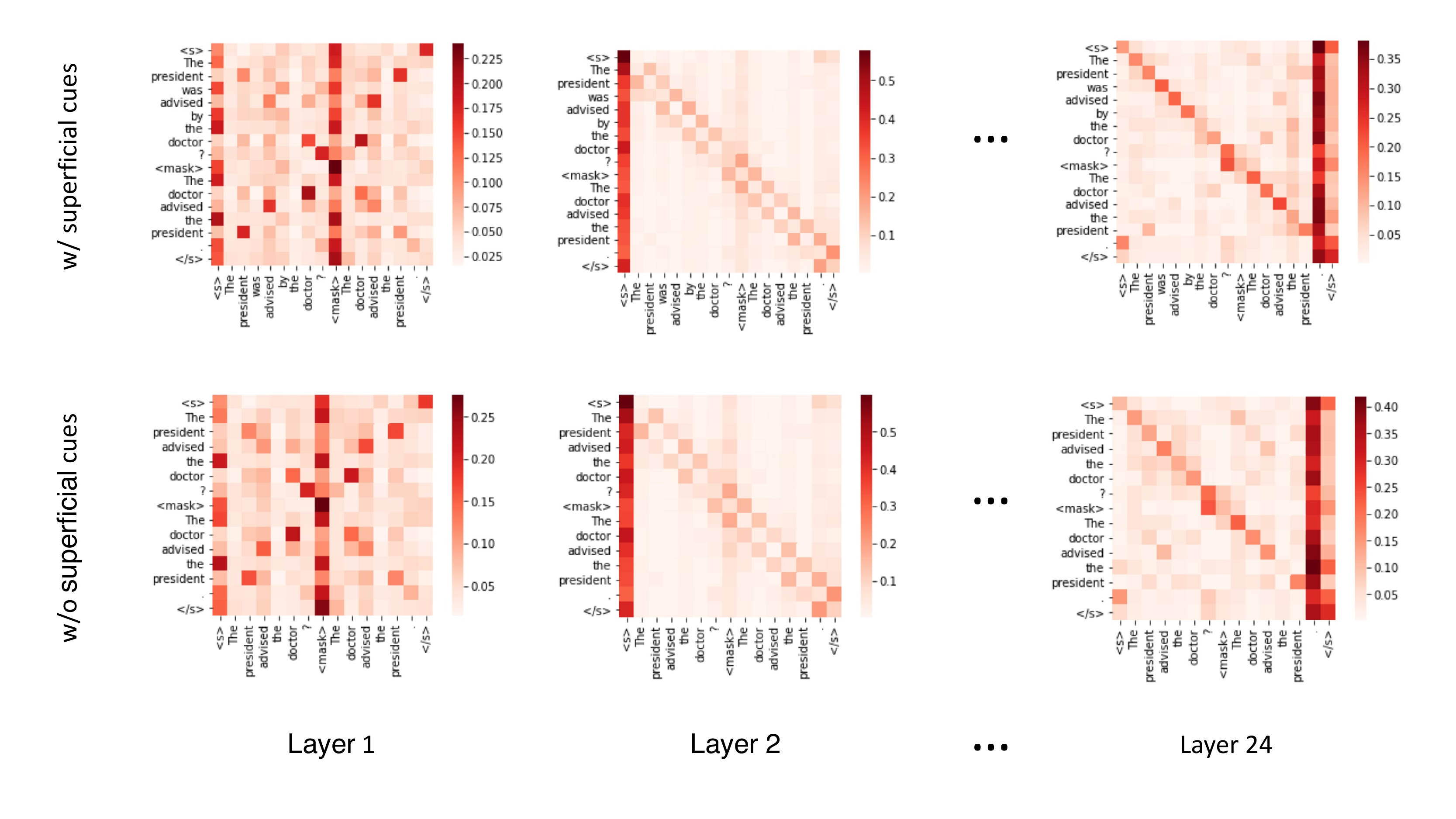}}

\caption{A comparison of prompt-based model attention maps on two natural language inference instances that have some words in common but have entirely different meanings.
The first row attention maps are for an instance with superficial cues, and the second row attention maps are for an instance without superficial cues (Appendix~\ref{app:attention} shows all layers).
}
\label{fig:attention-maps-hans}
\end{figure*}

\subsection{Sensitivity to Word Order in NLI}
\label{sec:meaning-usage-analysis}

The visual attention analysis revealed that the model does not respond well to change in meaning.
To investigate this at scale, we evaluate a model trained on input with correct word order and input with randomly shuffled word order.
We will refer to input with correct word order as \emph{meaningful} input and input with shuffled word order as \emph{meaningless} input.
Specifically, given an original test instance, we make it meaningless by randomly shuffling all the words in the instance while maintaining the English end of sentence punctuation mark if it exists in the original instance.
We do this so we can preserve the same number of English sentences as the original instance while making them meaningless.

For example, given the original NLI instance:

\nli
{The president was advised by the doctor.}
{The doctor advised the president.}
{Entailment}
The new instance becomes:
\nli
{The doctor by the president advised was.}
{The doctor the president advised.}
{Entailment}

If the model is sensitive to meaning, we expect the performance on this meaningless input to drop to random performance because the model was trained on meaningful input.

\begin{figure*}[t!]
    \centering
    \begin{subfigure}[t]{0.5\textwidth}
        \centering
        \includegraphics[width=\textwidth]{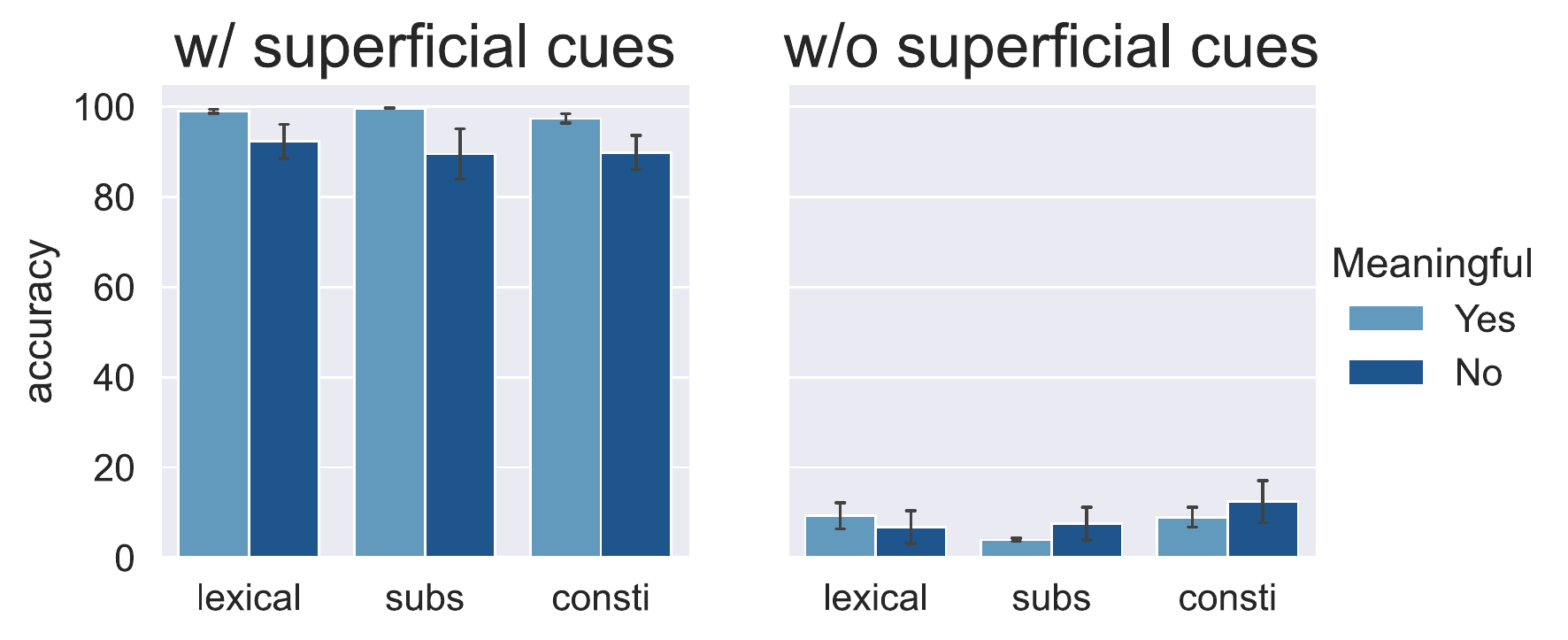}
        \caption{MNLI}
        \label{fig:mnli-hans-and-meaning}
    \end{subfigure}%
    ~ 
    \begin{subfigure}[t]{0.5\textwidth}
        \centering
        \includegraphics[width=\textwidth]{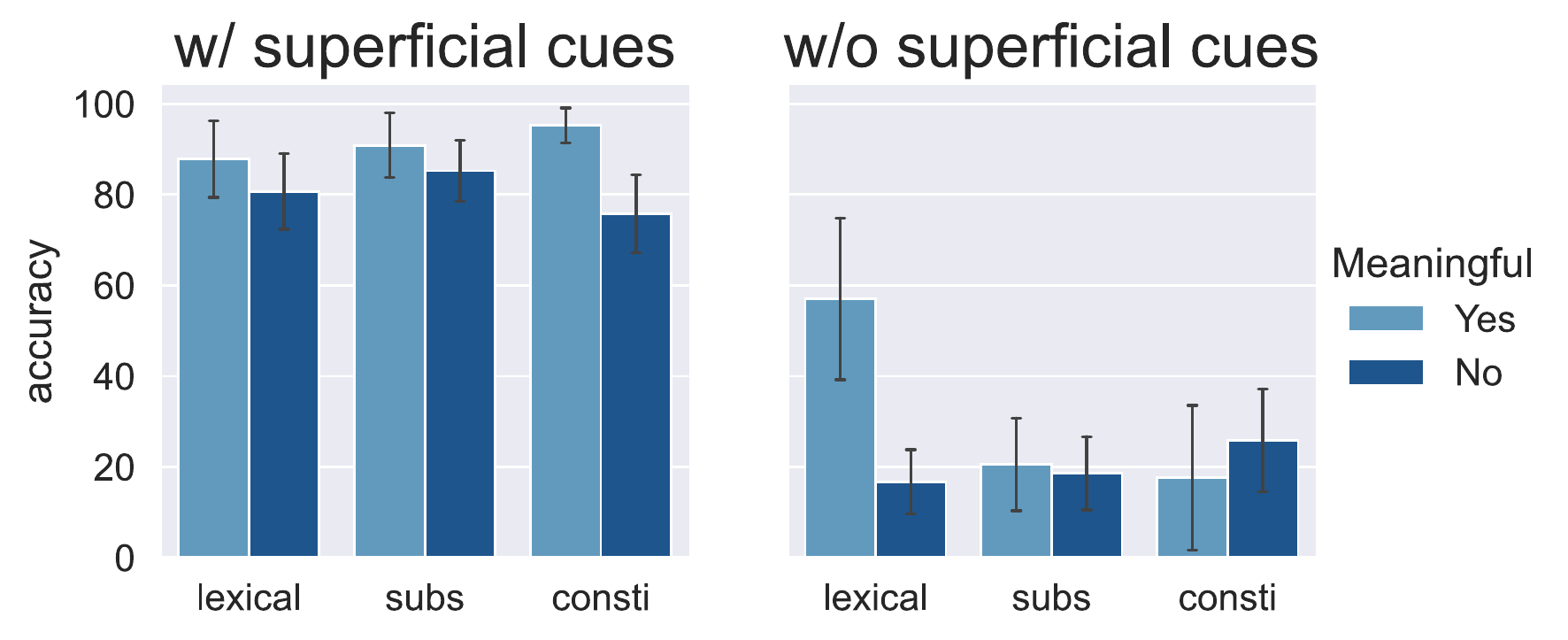}
        \caption{SNLI}
        \label{fig:snli-hans-and-meaning}
    \end{subfigure}
    \caption{The average accuracy of prompt-based RoBERTa trained on MNLI (a) and SNLI (b) instances with the correct word order (``Meaningful'') when evaluated on HANS dataset when the input has the correct word order (Yes in legend) and when the word order is shuffled (No in legend).
    }
    \label{fig:snli-mnli-hans-and-meaning}
\end{figure*}

Figure~\ref{fig:mnli-hans-and-meaning} and Figure~\ref{fig:snli-hans-and-meaning} shows the results of prompt-based RoBERTa trained on MNLI and SNLI, respectively.
The figures show the results of a prompt-based model trained on meaning-containing instances evaluated on the test set of instances whose meaning is preserved (Yes) and when the instances are made meaningless (No).
The results show that when meaning is removed from the instances, the model barely changes its predictions, indicating that the model hardly relies on the meaning of the instances.

\paragraph{Future Work} At this point, some questions still remain unanswered: (1) What are the specific superficial cues that the models exploit?
This still remains a hard interpretability question. 
(2) Are there any prompts that discourage models from exploiting superficial cues? (3) Can incorporating task demonstrations without superficial cues mitigate against the reliance on superficial cues?

\section{Related Works}
\label{sec:related-works}

\paragraph{Few-shot Prompting} 
Language model prompting was popularized by recent work of GPT-3~\cite{gpt3_NEURIPS2020_1457c0d6}.
\citet{gpt3_NEURIPS2020_1457c0d6} showed that by using prompts and some task demonstrations, GPT-3 could perform a number of tasks in the few shot setup.
Following this work, \citet{schick-schutze-2021-exploiting} showed that even much smaller language models such as RoBERTa-large could perform well when finetuned with prompts.
The subsequent work~\cite{schick-schutze-2021-just} demonstrated that a smaller model could achieve similar performance to GPT-3 in a few shot setup once finetuned with prompts.
Many works proposed better ways of generating prompts and answer keys~\cite{gao-etal-2021-making-lm-bff, le-scao-rush-2021-many-data-points}.
Our work develops on these works to gain more insights into model predictions.
Specifically, we undertake the first investigation of whether prompt-based models exploit superficial cues.

\paragraph{Superficial Cues in Datasets}
Superficial cues have been analyzed across several natural language understanding datasets.
\citet{gururangan-etal-2018-annotation} analyzed contextless superficial cues---hypothesis only superficial cues---in the MNLI dataset and SNLI dataset using fastText~\cite{fastText-joulin2017bag}, finding that a little over half of MNLI contain superficial cues and 67\% of SNLI contain superficial cues.
We argue that these figures could be outdated.
Hence, we reanalyze contextless superficial cues in MNLI and SNLI.
\citet{mccoy2019right} analyzed context superficial cues in MNLI, finding that lexical overlap is one superficial cue that can be exploited to correctly predict entailment labels.
Following their analysis, they released the HANS dataset that has an equal number of instances with superficial cues and those without superficial cues.
In this work, we use the HANS dataset to evaluate the models' ability to exploit superficial cues.
\citet{kavumba-etal-2019-choosing} analyzed contextless superficial cues in COPA using RoBERTa and productivity measures introduced by~\citet{niven2019probing}.
However, they did not analyze context superficial cues.
In this work, we analyze context superficial cues in COPA, and we use the contextless superficial cues from~\citet{kavumba-etal-2019-choosing}.

\paragraph{Exploiting Superficial Cues in Head Language Models}
Head language models that use a task-specific head for a downstream task have been analyzed on their ability to exploit superficial cues in datasets.
\citet{mccoy2019right} found that head BERT~\cite{devlin-etal-2019-bert} exploits superficial cues on MNLI dataset.
Similarly, \citet{niven2019probing} found that head BERT exploits superficial cues on the argument reasoning comprehension task and \citet{kavumba-etal-2019-choosing} analyzed BERT and RoBERTa's ability to exploit superficial cues on the COPA dataset.
While head models have been analyzed already, prompt-based models have not been analyzed yet. 
In this paper, we investigated whether prompt-based models also exploit superficial cues.

\section{Conclusions}
\label{sec:conclusion}

We presented the first analysis of whether prompt-based models exploit superficial cues.
We found that prompt-based models exploit superficial cues and fail to generalize well to instances without superficial cues on MNLI, SNLI, COPA, and HANS.
We, further, proposed to divide superficial cues into two: context and contextless superficial cues.
Our analysis of MNLI, SNLI, and COPA has revealed more superficial cues than was previously known.

\bibliography{anthology,custom}

\begin{thebibliography}{27}
\expandafter\ifx\csname natexlab\endcsname\relax\def\natexlab#1{#1}\fi

\bibitem[{Bowman et~al.(2015)Bowman, Angeli, Potts, and
  Manning}]{bowman-etal-2015-large-snli}
Samuel~R. Bowman, Gabor Angeli, Christopher Potts, and Christopher~D. Manning.
  2015.
\newblock \href {https://doi.org/10.18653/v1/D15-1075} {A large annotated
  corpus for learning natural language inference}.
\newblock In \emph{Proceedings of the 2015 Conference on Empirical Methods in
  Natural Language Processing}, pages 632--642, Lisbon, Portugal. Association
  for Computational Linguistics.

\bibitem[{Bras et~al.(2020)Bras, Swayamdipta, Bhagavatula, Zellers, Peters,
  Sabharwal, and Choi}]{pmlr-v119-bras20a-aflite}
Ronan~Le Bras, Swabha Swayamdipta, Chandra Bhagavatula, Rowan Zellers, Matthew
  Peters, Ashish Sabharwal, and Yejin Choi. 2020.
\newblock \href {https://proceedings.mlr.press/v119/bras20a.html} {Adversarial
  filters of dataset biases}.
\newblock In \emph{Proceedings of the 37th International Conference on Machine
  Learning}, volume 119 of \emph{Proceedings of Machine Learning Research},
  pages 1078--1088. PMLR.

\bibitem[{Brown et~al.(2020)Brown, Mann, Ryder, Subbiah, Kaplan, Dhariwal,
  Neelakantan, Shyam, Sastry, Askell, Agarwal, Herbert-Voss, Krueger, Henighan,
  Child, Ramesh, Ziegler, Wu, Winter, Hesse, Chen, Sigler, Litwin, Gray, Chess,
  Clark, Berner, McCandlish, Radford, Sutskever, and
  Amodei}]{gpt3_NEURIPS2020_1457c0d6}
Tom Brown, Benjamin Mann, Nick Ryder, Melanie Subbiah, Jared~D Kaplan, Prafulla
  Dhariwal, Arvind Neelakantan, Pranav Shyam, Girish Sastry, Amanda Askell,
  Sandhini Agarwal, Ariel Herbert-Voss, Gretchen Krueger, Tom Henighan, Rewon
  Child, Aditya Ramesh, Daniel Ziegler, Jeffrey Wu, Clemens Winter, Chris
  Hesse, Mark Chen, Eric Sigler, Mateusz Litwin, Scott Gray, Benjamin Chess,
  Jack Clark, Christopher Berner, Sam McCandlish, Alec Radford, Ilya Sutskever,
  and Dario Amodei. 2020.
\newblock \href
  {https://proceedings.neurips.cc/paper/2020/file/1457c0d6bfcb4967418bfb8ac142f64a-Paper.pdf}
  {Language models are few-shot learners}.
\newblock In \emph{Advances in Neural Information Processing Systems},
  volume~33, pages 1877--1901. Curran Associates, Inc.

\bibitem[{Devlin et~al.(2019)Devlin, Chang, Lee, and
  Toutanova}]{devlin-etal-2019-bert}
Jacob Devlin, Ming-Wei Chang, Kenton Lee, and Kristina Toutanova. 2019.
\newblock \href {https://doi.org/10.18653/v1/N19-1423} {{BERT}: Pre-training of
  deep bidirectional transformers for language understanding}.
\newblock In \emph{Proceedings of the 2019 Conference of the North {A}merican
  Chapter of the Association for Computational Linguistics: Human Language
  Technologies, Volume 1 (Long and Short Papers)}, pages 4171--4186,
  Minneapolis, Minnesota. Association for Computational Linguistics.

\bibitem[{Gao et~al.(2021)Gao, Fisch, and Chen}]{gao-etal-2021-making-lm-bff}
Tianyu Gao, Adam Fisch, and Danqi Chen. 2021.
\newblock \href {https://doi.org/10.18653/v1/2021.acl-long.295} {Making
  pre-trained language models better few-shot learners}.
\newblock In \emph{Proceedings of the 59th Annual Meeting of the Association
  for Computational Linguistics and the 11th International Joint Conference on
  Natural Language Processing (Volume 1: Long Papers)}, pages 3816--3830,
  Online. Association for Computational Linguistics.

\bibitem[{Gururangan et~al.(2018)Gururangan, Swayamdipta, Levy, Schwartz,
  Bowman, and Smith}]{gururangan-etal-2018-annotation}
Suchin Gururangan, Swabha Swayamdipta, Omer Levy, Roy Schwartz, Samuel Bowman,
  and Noah~A. Smith. 2018.
\newblock \href {https://doi.org/10.18653/v1/N18-2017} {Annotation artifacts in
  natural language inference data}.
\newblock In \emph{Proceedings of the 2018 Conference of the North {A}merican
  Chapter of the Association for Computational Linguistics: Human Language
  Technologies, Volume 2 (Short Papers)}, pages 107--112, New Orleans,
  Louisiana. Association for Computational Linguistics.

\bibitem[{Habernal et~al.(2018)Habernal, Wachsmuth, Gurevych, and
  Stein}]{habernal-etal-2018-semeval-arct}
Ivan Habernal, Henning Wachsmuth, Iryna Gurevych, and Benno Stein. 2018.
\newblock \href {https://doi.org/10.18653/v1/S18-1121} {{S}em{E}val-2018 task
  12: The argument reasoning comprehension task}.
\newblock In \emph{Proceedings of The 12th International Workshop on Semantic
  Evaluation}, pages 763--772, New Orleans, Louisiana. Association for
  Computational Linguistics.

\bibitem[{Joulin et~al.(2017)Joulin, Grave, Bojanowski, and
  Mikolov}]{fastText-joulin2017bag}
Armand Joulin, Edouard Grave, Piotr Bojanowski, and Tomas Mikolov. 2017.
\newblock Bag of tricks for efficient text classification.
\newblock In \emph{Proceedings of the 15th Conference of the European Chapter
  of the Association for Computational Linguistics: Volume 2, Short Papers},
  pages 427--431. Association for Computational Linguistics.

\bibitem[{Kavumba et~al.(2019)Kavumba, Inoue, Heinzerling, Singh, Reisert, and
  Inui}]{kavumba-etal-2019-choosing}
Pride Kavumba, Naoya Inoue, Benjamin Heinzerling, Keshav Singh, Paul Reisert,
  and Kentaro Inui. 2019.
\newblock \href {https://doi.org/10.18653/v1/D19-6004} {When choosing plausible
  alternatives, clever hans can be clever}.
\newblock In \emph{Proceedings of the First Workshop on Commonsense Inference
  in Natural Language Processing}, pages 33--42, Hong Kong, China. Association
  for Computational Linguistics.

\bibitem[{Le~Scao and Rush(2021)}]{le-scao-rush-2021-many-data-points}
Teven Le~Scao and Alexander Rush. 2021.
\newblock \href {https://doi.org/10.18653/v1/2021.naacl-main.208} {How many
  data points is a prompt worth?}
\newblock In \emph{Proceedings of the 2021 Conference of the North American
  Chapter of the Association for Computational Linguistics: Human Language
  Technologies}, pages 2627--2636, Online. Association for Computational
  Linguistics.

\bibitem[{Liu et~al.(2021)Liu, Yuan, Fu, Jiang, Hayashi, and
  Neubig}]{liu2021pretrain-prompt-survey}
Pengfei Liu, Weizhe Yuan, Jinlan Fu, Zhengbao Jiang, Hiroaki Hayashi, and
  Graham Neubig. 2021.
\newblock \href {http://arxiv.org/abs/2107.13586} {Pre-train, prompt, and
  predict: A systematic survey of prompting methods in natural language
  processing}.

\bibitem[{Liu et~al.(2019)Liu, Ott, Goyal, Du, Joshi, Chen, Levy, Lewis,
  Zettlemoyer, and Stoyanov}]{RoBERTa2019}
Yinhan Liu, Myle Ott, Naman Goyal, Jingfei Du, Mandar Joshi, Danqi Chen, Omer
  Levy, Mike Lewis, Luke Zettlemoyer, and Veselin Stoyanov. 2019.
\newblock \href {http://arxiv.org/abs/1907.11692} {Roberta: {A} robustly
  optimized {BERT} pretraining approach}.
\newblock \emph{CoRR}, abs/1907.11692.

\bibitem[{McCoy et~al.(2019)McCoy, Pavlick, and Linzen}]{mccoy2019right}
Tom McCoy, Ellie Pavlick, and Tal Linzen. 2019.
\newblock \href {https://www.aclweb.org/anthology/P19-1334} {Right for the
  wrong reasons: Diagnosing syntactic heuristics in natural language
  inference}.
\newblock In \emph{Proceedings of the 57th Annual Meeting of the Association
  for Computational Linguistics}, pages 3428--3448, Florence, Italy.
  Association for Computational Linguistics.

\bibitem[{Niven and Kao(2019)}]{niven2019probing}
Timothy Niven and Hung-Yu Kao. 2019.
\newblock \href {https://www.aclweb.org/anthology/P19-1459} {Probing neural
  network comprehension of natural language arguments}.
\newblock In \emph{Proceedings of the 57th Annual Meeting of the Association
  for Computational Linguistics}, pages 4658--4664, Florence, Italy.
  Association for Computational Linguistics.

\bibitem[{Poliak et~al.(2018)Poliak, Naradowsky, Haldar, Rudinger, and
  Van~Durme}]{poliak-etal-2018-hypothesis}
Adam Poliak, Jason Naradowsky, Aparajita Haldar, Rachel Rudinger, and Benjamin
  Van~Durme. 2018.
\newblock \href {https://doi.org/10.18653/v1/S18-2023} {Hypothesis only
  baselines in natural language inference}.
\newblock In \emph{Proceedings of the Seventh Joint Conference on Lexical and
  Computational Semantics}, pages 180--191, New Orleans, Louisiana. Association
  for Computational Linguistics.

\bibitem[{Roemmele et~al.(2011)Roemmele, Bejan, and
  Gordon}]{roemmele2011choice}
Melissa Roemmele, Cosmin~Adrian Bejan, and Andrew~S Gordon. 2011.
\newblock \href
  {http://people.ict.usc.edu/~gordon/publications/AAAI-SPRING11A.PDF} {Choice
  of plausible alternatives: An evaluation of commonsense causal reasoning}.
\newblock In \emph{AAAI Spring Symposium on Logical Formalizations of
  Commonsense Reasoning}, Stanford University.

\bibitem[{Schick and
  Sch{\"u}tze(2021{\natexlab{a}})}]{schick-schutze-2021-exploiting}
Timo Schick and Hinrich Sch{\"u}tze. 2021{\natexlab{a}}.
\newblock \href {https://aclanthology.org/2021.eacl-main.20} {Exploiting
  cloze-questions for few-shot text classification and natural language
  inference}.
\newblock In \emph{Proceedings of the 16th Conference of the European Chapter
  of the Association for Computational Linguistics: Main Volume}, pages
  255--269, Online. Association for Computational Linguistics.

\bibitem[{Schick and
  Sch{\"u}tze(2021{\natexlab{b}})}]{schick-schutze-2021-just}
Timo Schick and Hinrich Sch{\"u}tze. 2021{\natexlab{b}}.
\newblock \href {https://doi.org/10.18653/v1/2021.naacl-main.185} {It{'}s not
  just size that matters: Small language models are also few-shot learners}.
\newblock In \emph{Proceedings of the 2021 Conference of the North American
  Chapter of the Association for Computational Linguistics: Human Language
  Technologies}, pages 2339--2352, Online. Association for Computational
  Linguistics.

\bibitem[{Schuster et~al.(2019)Schuster, Shah, Yeo, Roberto Filizzola~Ortiz,
  Santus, and Barzilay}]{schuster-etal-2019-towards}
Tal Schuster, Darsh Shah, Yun Jie~Serene Yeo, Daniel Roberto Filizzola~Ortiz,
  Enrico Santus, and Regina Barzilay. 2019.
\newblock \href {https://doi.org/10.18653/v1/D19-1341} {Towards debiasing fact
  verification models}.
\newblock In \emph{Proceedings of the 2019 Conference on Empirical Methods in
  Natural Language Processing and the 9th International Joint Conference on
  Natural Language Processing (EMNLP-IJCNLP)}, pages 3410--3416, Hong Kong,
  China. Association for Computational Linguistics.

\bibitem[{Sugawara et~al.(2018)Sugawara, Inui, Sekine, and
  Aizawa}]{sugawara-etal-2018-makes}
Saku Sugawara, Kentaro Inui, Satoshi Sekine, and Akiko Aizawa. 2018.
\newblock \href {https://doi.org/10.18653/v1/D18-1453} {What makes reading
  comprehension questions easier?}
\newblock In \emph{Proceedings of the 2018 Conference on Empirical Methods in
  Natural Language Processing}, pages 4208--4219, Brussels, Belgium.
  Association for Computational Linguistics.

\bibitem[{Trichelair et~al.(2019)Trichelair, Emami, Trischler, Suleman, and
  Cheung}]{trichelair-etal-2019-reasonable}
Paul Trichelair, Ali Emami, Adam Trischler, Kaheer Suleman, and Jackie Chi~Kit
  Cheung. 2019.
\newblock \href {https://doi.org/10.18653/v1/D19-1335} {How reasonable are
  common-sense reasoning tasks: A case-study on the {W}inograd schema challenge
  and {SWAG}}.
\newblock In \emph{Proceedings of the 2019 Conference on Empirical Methods in
  Natural Language Processing and the 9th International Joint Conference on
  Natural Language Processing (EMNLP-IJCNLP)}, pages 3373--3378, Hong Kong,
  China. Association for Computational Linguistics.

\bibitem[{Vig and Belinkov(2019)}]{vig-belinkov-2019-analyzing-attention}
Jesse Vig and Yonatan Belinkov. 2019.
\newblock \href {https://doi.org/10.18653/v1/W19-4808} {Analyzing the structure
  of attention in a transformer language model}.
\newblock In \emph{Proceedings of the 2019 ACL Workshop BlackboxNLP: Analyzing
  and Interpreting Neural Networks for NLP}, pages 63--76, Florence, Italy.
  Association for Computational Linguistics.

\bibitem[{Wang et~al.(2019)Wang, Pruksachatkun, Nangia, Singh, Michael, Hill,
  Levy, and Bowman}]{wang2019superglue}
Alex Wang, Yada Pruksachatkun, Nikita Nangia, Amanpreet Singh, Julian Michael,
  Felix Hill, Omer Levy, and Samuel~R. Bowman. 2019.
\newblock Super{GLUE}: A stickier benchmark for general-purpose language
  understanding systems.
\newblock \emph{arXiv preprint 1905.00537}.

\bibitem[{Wang et~al.(2018)Wang, Singh, Michael, Hill, Levy, and
  Bowman}]{wang-etal-2018-glue}
Alex Wang, Amanpreet Singh, Julian Michael, Felix Hill, Omer Levy, and Samuel
  Bowman. 2018.
\newblock \href {https://doi.org/10.18653/v1/W18-5446} {{GLUE}: A multi-task
  benchmark and analysis platform for natural language understanding}.
\newblock In \emph{Proceedings of the 2018 {EMNLP} Workshop {B}lackbox{NLP}:
  Analyzing and Interpreting Neural Networks for {NLP}}, pages 353--355,
  Brussels, Belgium. Association for Computational Linguistics.

\bibitem[{Wiegreffe and
  Pinter(2019)}]{wiegreffe-pinter-2019-attention-not-explanation}
Sarah Wiegreffe and Yuval Pinter. 2019.
\newblock \href {https://doi.org/10.18653/v1/D19-1002} {Attention is not not
  explanation}.
\newblock In \emph{Proceedings of the 2019 Conference on Empirical Methods in
  Natural Language Processing and the 9th International Joint Conference on
  Natural Language Processing (EMNLP-IJCNLP)}, pages 11--20, Hong Kong, China.
  Association for Computational Linguistics.

\bibitem[{Williams et~al.(2018)Williams, Nangia, and
  Bowman}]{williams-etal-2018-broad}
Adina Williams, Nikita Nangia, and Samuel Bowman. 2018.
\newblock \href {https://doi.org/10.18653/v1/N18-1101} {A broad-coverage
  challenge corpus for sentence understanding through inference}.
\newblock In \emph{Proceedings of the 2018 Conference of the North {A}merican
  Chapter of the Association for Computational Linguistics: Human Language
  Technologies, Volume 1 (Long Papers)}, pages 1112--1122, New Orleans,
  Louisiana. Association for Computational Linguistics.

\bibitem[{Wolf et~al.(2019)Wolf, Debut, Sanh, Chaumond, Delangue, Moi, Cistac,
  Rault, Louf, Funtowicz, and Brew}]{Wolf2019HuggingFacesTS}
Thomas Wolf, Lysandre Debut, Victor Sanh, Julien Chaumond, Clement Delangue,
  Anthony Moi, Pierric Cistac, Tim Rault, R'emi Louf, Morgan Funtowicz, and
  Jamie Brew. 2019.
\newblock Huggingface's transformers: State-of-the-art natural language
  processing.
\newblock \emph{ArXiv}, abs/1910.03771.

\end{thebibliography}
\bibliographystyle{acl_natbib}

\clearpage
\appendix

\section{Datasets}
\label{app:datasets}

\subsection{MNLI}

The Multi-Genre Natural Language Inference~\cite[MNLI]{williams-etal-2018-broad} dataset is an important dataset of natural language inference which is also part of the SuperGLUE benchmark~\cite{wang2019superglue}. Given a premise, such as \emph{``The Old One always comforted Ca'daan, except today.''}, and a hypothesis, in this case, \emph{``Ca'daan knew the Old One very well.''} (neutral), a model is asked to pick one label from among three, \{contradiction,  neutral, entailment\}.
The test set of MNLI is divided into two subsets according to whether the domain of each test instance matches the domain of the training set: matched (in-domain instances) and mismatched (out-of-domain instances).

\subsection{SNLI}

The Stanford Natural Language Inference~\cite[SNLI]{bowman-etal-2015-large-snli} is a popular natural language inference dataset with the same format as MNLI.

\subsection{HANS Datasets}
\citet{mccoy2019right} identified three context superficial cues in MNLI: \overlap{}, \subs{}, and \const{}.
These superficial cues are predictive of entailment label.
Both \subs{} and \const{} superficial cues are special cases of lexical \overlap{}.
Following this analysis, \citet{mccoy2019right} created the HANS (Heuristic Analysis for NLI Systems) dataset, which includes instances with superficial cues---lexical overlap entails an entailment label and instances without superficial cues---lexical overlap does not necessarily mean it is an entailment label.

\subsection{COPA} 

The Choice of Plausible Alternatives~\cite[COPA]{roemmele2011choice} dataset is a popular commonsense dataset that asks the model to select the most plausible answer choice from the set of two candidates, which is also a part of the SuperGLUE benchmark.
For example, given a premise such as \emph{``I tipped the bottle.''} and a question that can either be the \emph{cause} or \emph{effect}, the model is asked to select the most plausible alternative from either \emph{``The liquid in the bottle froze.''} or \emph{``The liquid in the bottle poured out.''} (correct).

\section{Prompt Templates and Answer Keys}
\label{app:prompt-templates}

In the templates we show the part that is replaced with actual question input between two braces (\{\}), and we represent the mask token for a language model with <MASK>.

\subsection{MNLI}

We use the template and answer keys from~\cite{gao-etal-2021-making-lm-bff}.

\noindent
Template:

\begin{center}
\texttt{\{premise\}? <MASK>, \{hypothesis\}}
\end{center}

\noindent
For the answer key that maps from model output to task labels we used:

\begin{center}
\texttt{\{yes: entailment, maybe: netural, No: contradiction\}}
\end{center}

\subsection{SNLI}

We use the template and answer keys from~\cite{gao-etal-2021-making-lm-bff}.

\noindent
Template:

\begin{center}
\texttt{\{premise\}? <MASK>, \{hypothesis\}}
\end{center}

\noindent
For the \emph{answer key} that maps from model output to task labels we used:

\begin{center}
\texttt{\{yes: entailment, maybe: netural, No: contradiction\}}
\end{center}

\subsection{COPA}

For COPA we use different templates from~\citet{schick-schutze-2021-just, le-scao-rush-2021-many-data-points} depending on the question type.
COPA instances can have a \emph{cause} or \emph{effect} question.

For a \emph{cause} question with a given \emph{premise} and two possible causes; \emph{choice1} and \emph{choice2}. We use the template;

\begin{table*}[t]
\centering
\begin{tabular}{@{}llrrrr@{}}
\toprule
\multirow{2}{*}{Train set} & \multirow{2}{*}{Cue Type} & \multicolumn{2}{c}{W/ Superficial Cues}                          & \multicolumn{2}{c}{W/O Superficial Cues}                         \\ \cmidrule(l){3-6} 
                           &                           & \multicolumn{1}{c}{Meaningful} & \multicolumn{1}{c}{Meaningless} & \multicolumn{1}{c}{Meaningful} & \multicolumn{1}{c}{Meaningless} \\ \midrule %
\multirow{3}{*}{MNLI}      & lexical\_overlap           & 99.0 \small{$\pm$0.4}          & 92.3\small{$\pm$3.7}           & 9.3 \small{$\pm$2.9}           & 6.8 \small{$\pm$3.7}            \\
                           & subsequence               & 99.6 \small{$\pm$0.1}          & 89.5 \small{$\pm$5.6}           & 3.9 \small{$\pm$0.3}           & 7.6 \small{$\pm$3.6}            \\
                           & constituent               & 97.4 \small{$\pm$1.0}          & 89.9 \small{$\pm$3.8}           & 8.9 \small{$\pm$2.2}           & 12.4 \small{$\pm$4.7}           \\ \cmidrule(l){1-6}
\multirow{3}{*}{SNLI}      & lexical\_overlap           & 87.8 \small{$\pm$8.4}          & 80.7 \small{$\pm$8.3}           & 57.0 \small{$\pm$17.8}         & 16.7 \small{$\pm$7.1}           \\
                           & subsequence               & 90.9 \small{$\pm$7.1}          & 85.2 \small{$\pm$6.7}           & 20.6 \small{$\pm$10.2}         & 18.6 \small{$\pm$8.1}           \\
                           & constituent               & 95.3 \small{$\pm$3.9}          & 75.8 \small{$\pm$8.6}           & 17.6 \small{$\pm$15.9}         & 25.8 \small{$\pm$11.3}          \\ \bottomrule %
\end{tabular}
\caption{Average accuracy on HANS Datasets for prompt-based RoBERTa models trained on meaningful MNLI and SNLI; and evaluated on instances whose meaning has been preserved (Meaningful) and instances whose meaning has been removed (Meaningless). The instances are further divided into instances with (W/) and without (W/O) context superficial cues}
\label{app-tab:hans-meaning-meaningless}
\end{table*}
\begin{table}[t]
\centering
\begin{tabular}{@{}llll@{}}
\toprule
Dataset & w/ s. cues                  & w/o s. cues                  & Overall               \\ \midrule

MNLI    & 69.5{\small $\pm$0.1} & 38.6{\small $\pm$0.6} & 66.7{\small $\pm$0.1} \\
MNLI-mm & 72.0{\small $\pm$0.2} & 41.5{\small $\pm$0.7} & 69.1{\small $\pm$0.2} \\ \cmidrule(l){1-4}
SNLI    & 82.2{\small $\pm$0.9} & 67.1{\small $\pm$1.0} & 78.0{\small $\pm$0.9} \\
\bottomrule
\end{tabular}
\caption{Results of of prompt-based RoBERTa on MNLI and SNLI instances with contextless superficial cues in the hypothesis and instances without contextless superficial cues.}
\label{app-tab:mnli-snli-contextless}
\end{table}

\begin{center}
\texttt{``\{choice1\}'' or ``\{choice2\}''? \{premise\}, so <MASK>.}
\end{center}

\noindent
and 

\begin{center}
\texttt{ \{choice1\} or \{choice2\}? \{premise\}, so <MASK>.}
\end{center}

\noindent
For an \emph{effect} question with a given \emph{premise} and two possible causes; \emph{choice1} and \emph{choice2}. We use the template;

\begin{center}
\texttt{``\{choice1\}'' or ``\{choice2\}''? <MASK>, because \{premise\}.}
\end{center}

\noindent
and

\begin{center}
\texttt{\quad \quad \{choice1\} or \{choice2\}? <MASK>, because \{premise\}.}
\end{center}

For the answer key we use the identity function.

\section{Training Details}
\label{app:training-details}
All our experiments are run on a single NVIDIA Tesla T4 GPU with 16GB memory.
All the models are trained in a few-shot setup.
Thus they training only takes a few minutes to complete.
We use a learning rate of 1e-5 for all our experiments.

\section{Numeric Results}
\label{app:results-in-tables}
For all the figures containing results---figure 2 and figure 4---we also include numeric results in table~\ref{app-tab:hans-meaning-meaningless} and table~\ref{app-tab:mnli-snli-contextless}---to show the exact values and standard deviations to make it easier to compare with future work.

\section{Attention Maps}
\label{app:attention}
This appendix shows the complete attention maps partially visualized in section~\ref{sec:attention-analysis}.

The attention maps are for a instance with superficial cues below.
\nli
{The president was advised by the doctor.}
{The doctor advised the president.}
{Entailment}

And an instance without superficial cues:
\nli
{The president advised the doctor.}
{The doctor advised the president.}
{Non-Entailment}

To make it easier to compare the attention maps, we alternate the pages between the attention maps for an instance with and without superficial cues.
Each page has six attention maps to make it easier to see all the points.

\begin{figure*}[t]
\includegraphics[width=\textwidth]{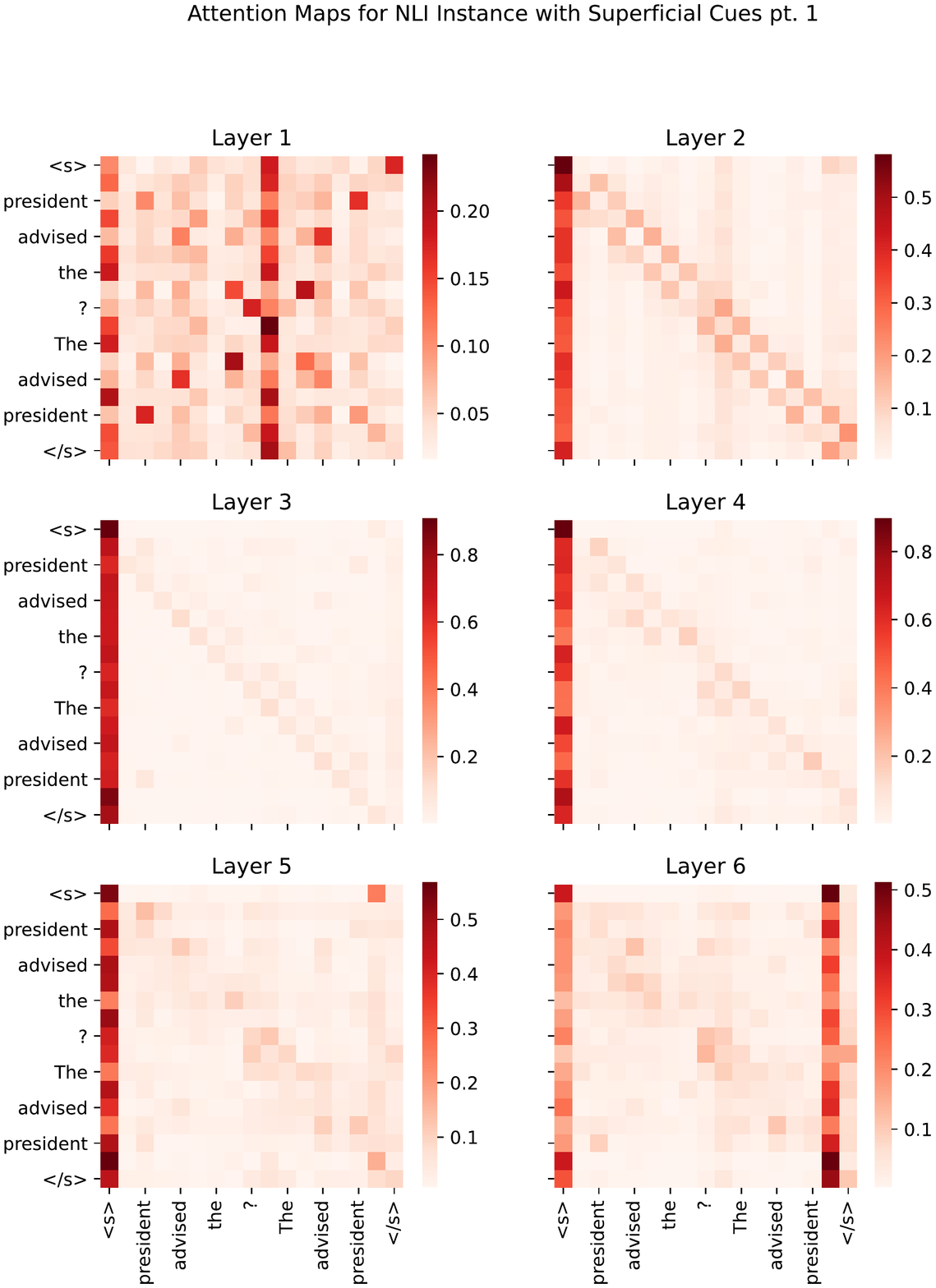}
\end{figure*}

\begin{figure*}[t]
\includegraphics[width=\textwidth]{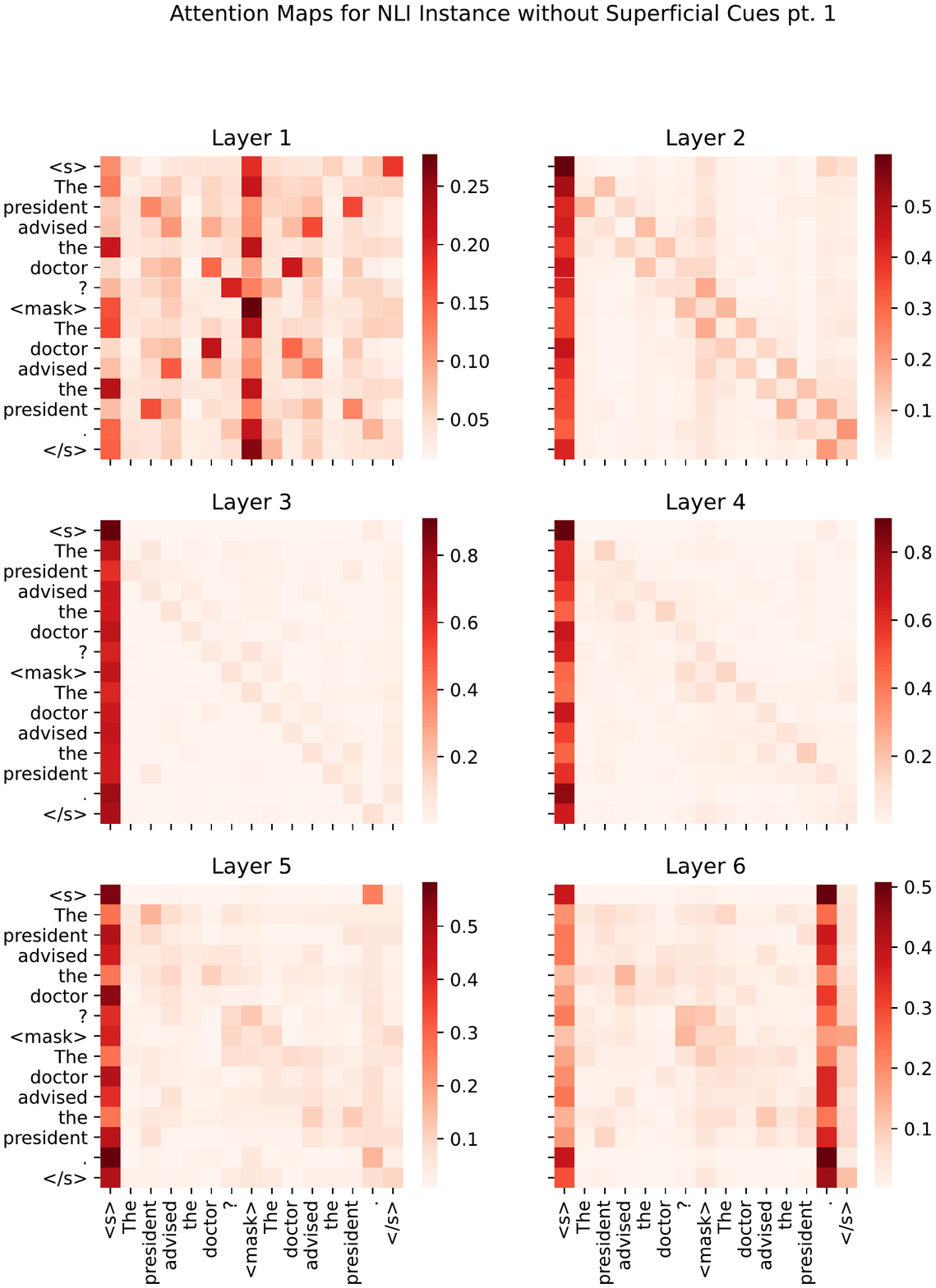}
\end{figure*}

\begin{figure*}[t]
\includegraphics[width=\textwidth]{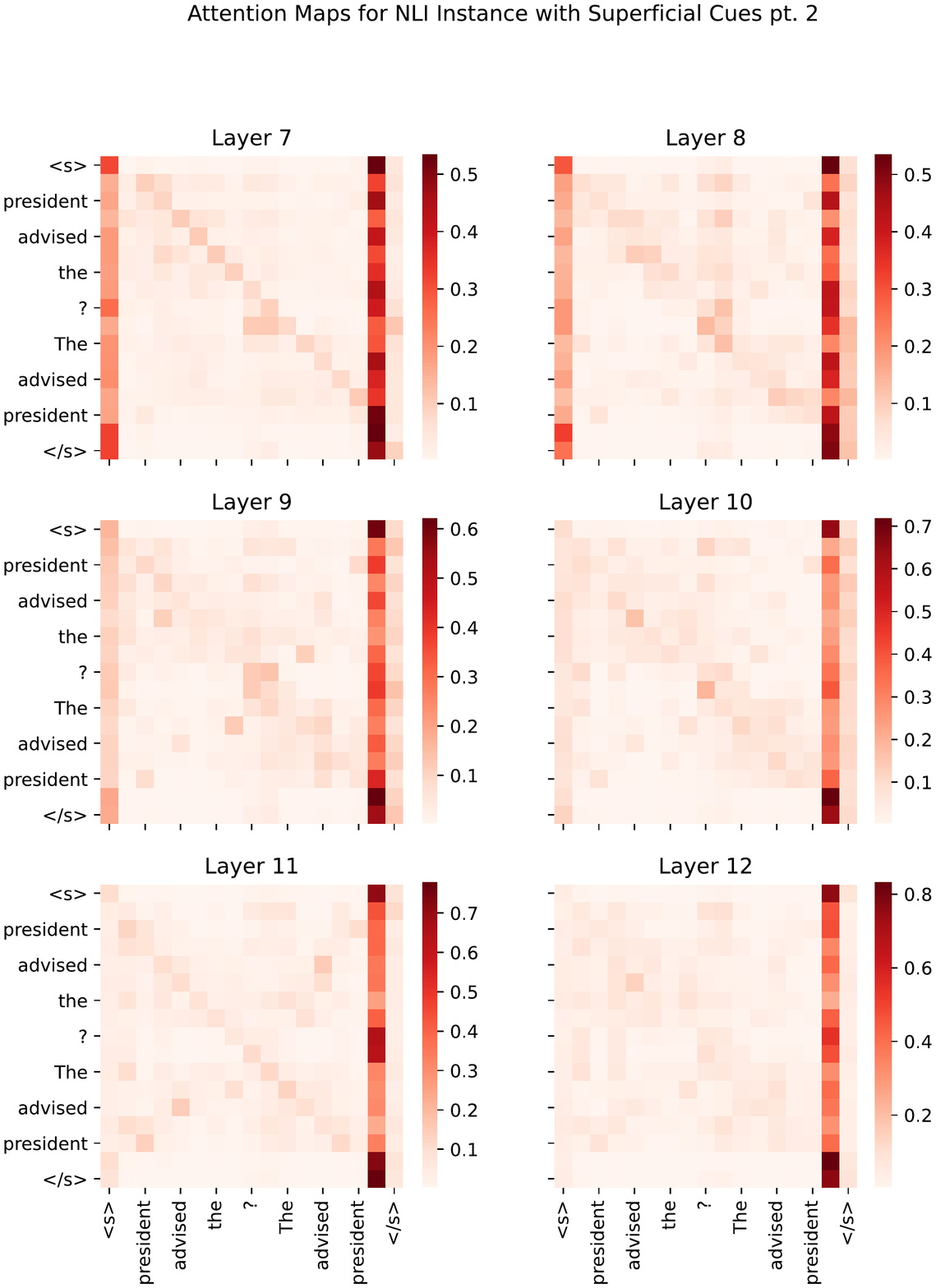}
\end{figure*}
\begin{figure*}[t]
\includegraphics[width=\textwidth]{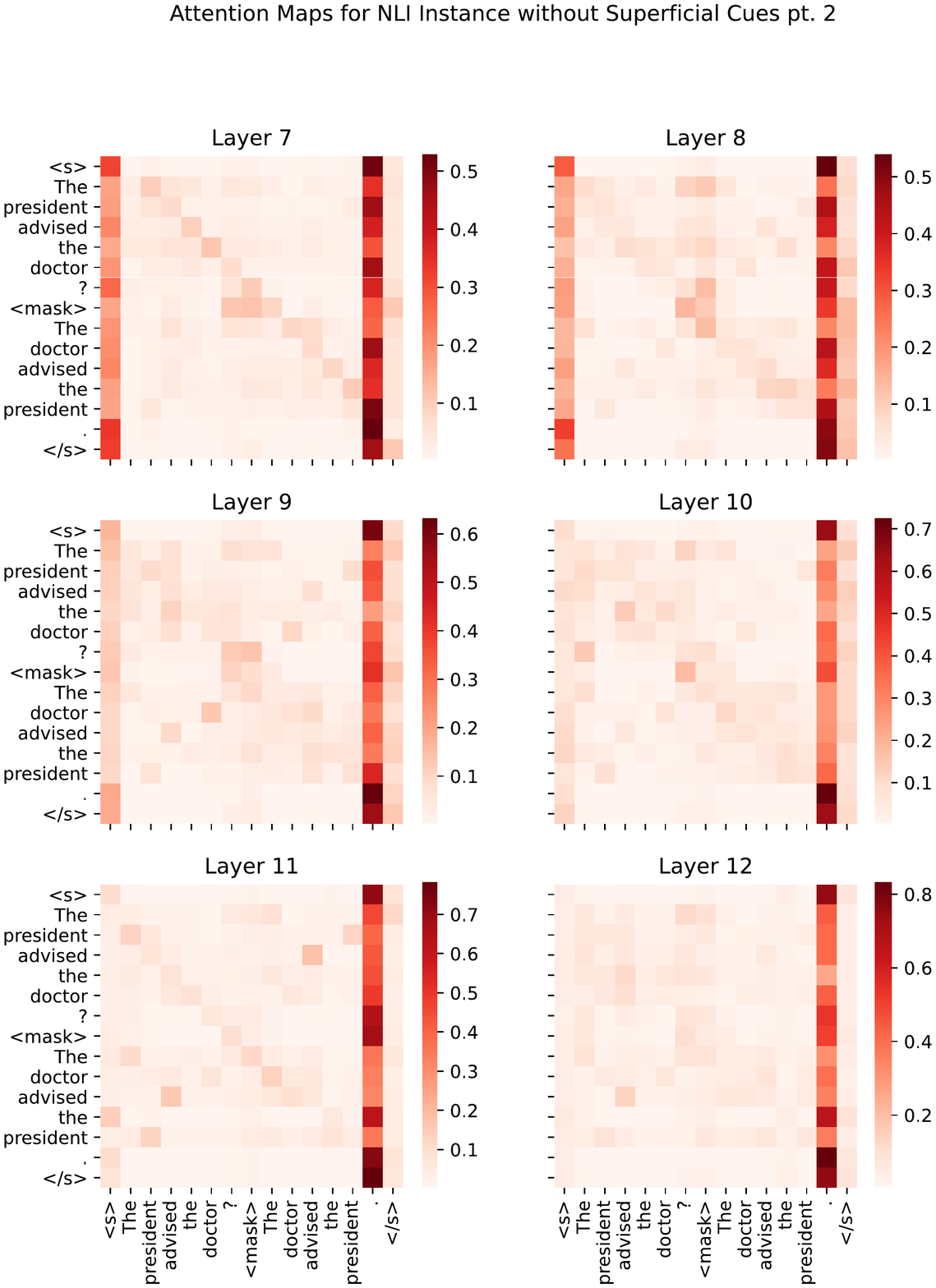}
\end{figure*}

\begin{figure*}[t]
\includegraphics[width=\textwidth]{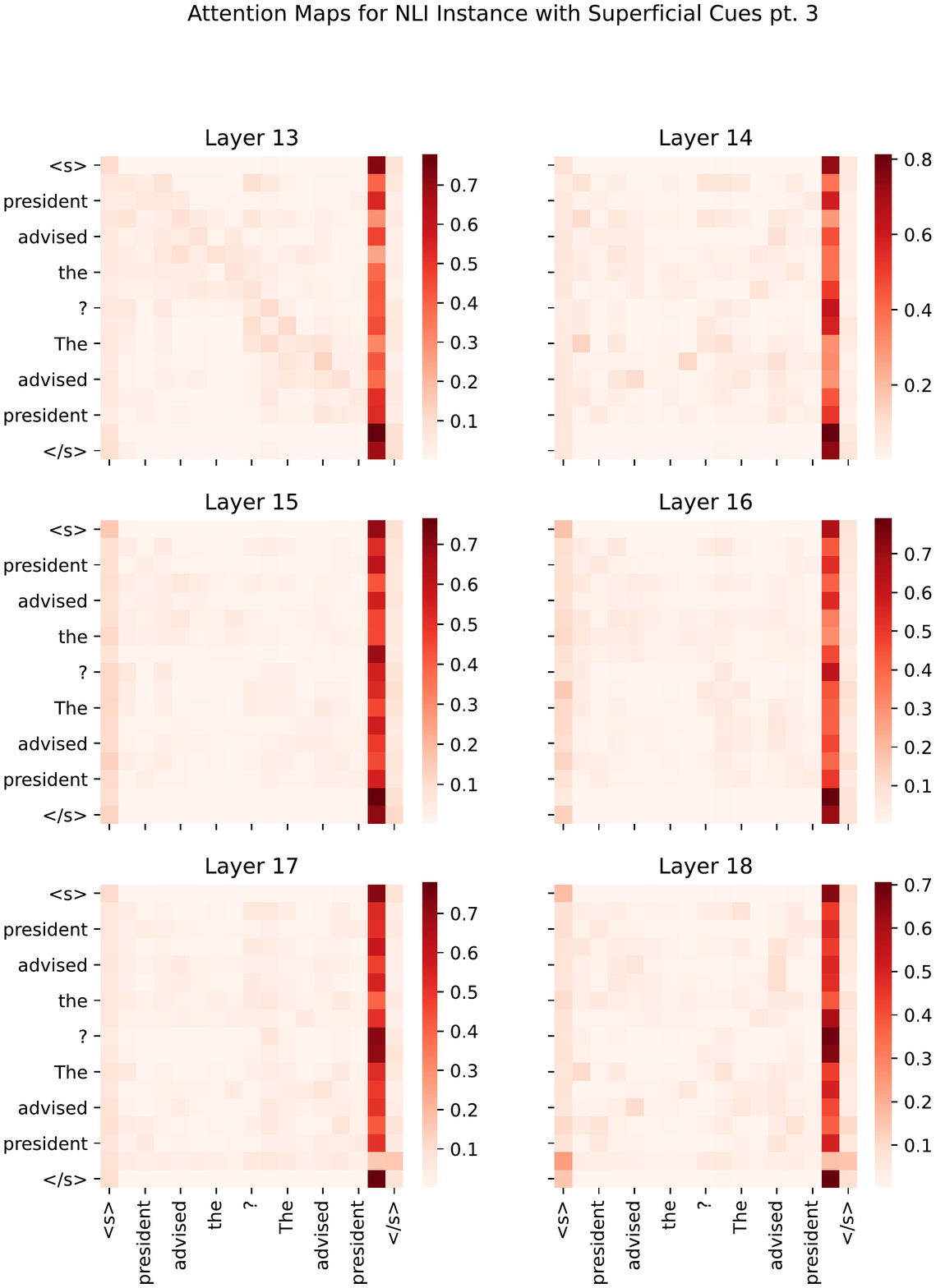}
\end{figure*}
\begin{figure*}[t]
\includegraphics[width=\textwidth]{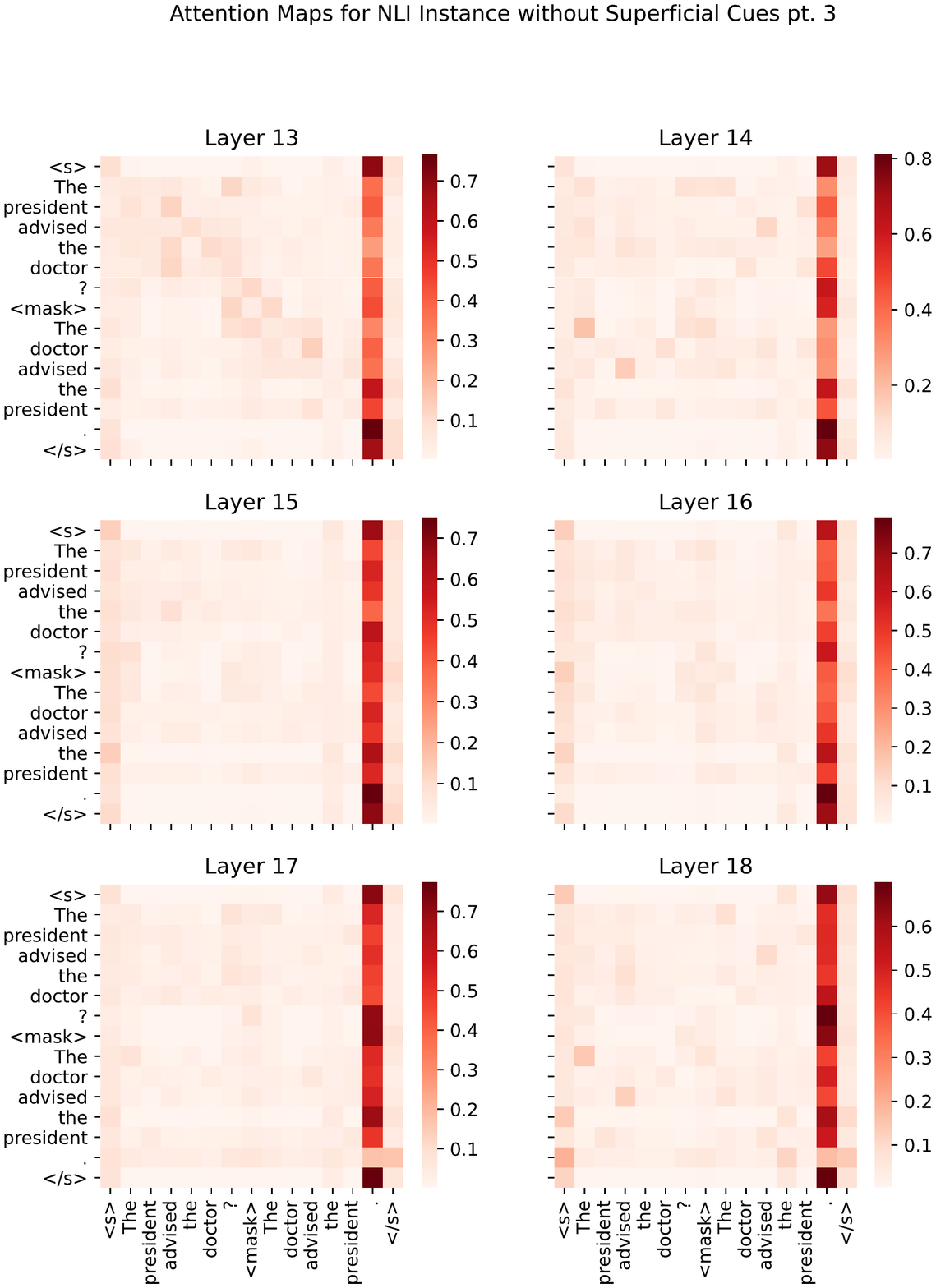}
\end{figure*}

\begin{figure*}[t]
\includegraphics[width=\textwidth]{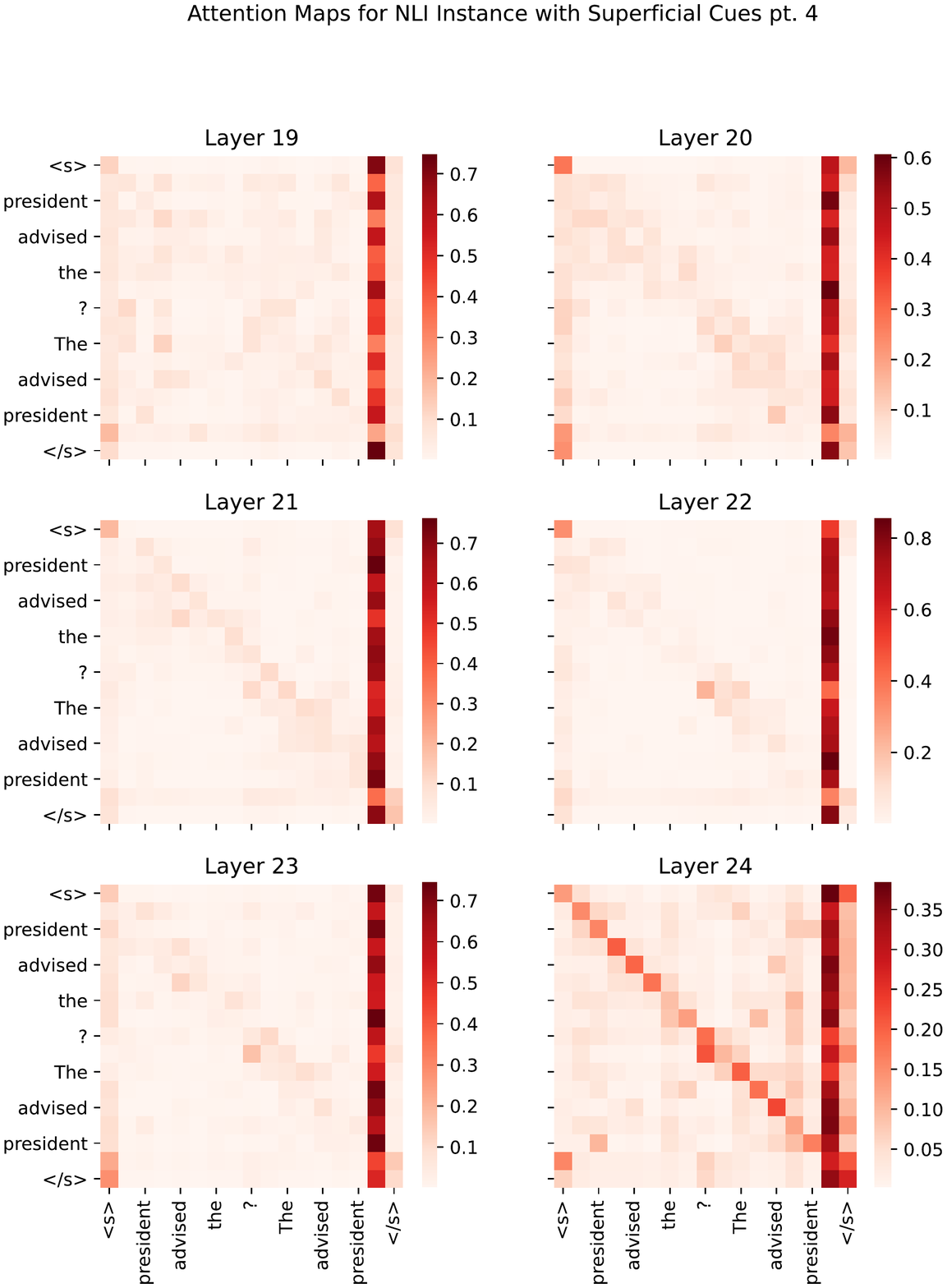}
\end{figure*}
\begin{figure*}[t]
\includegraphics[width=\textwidth]{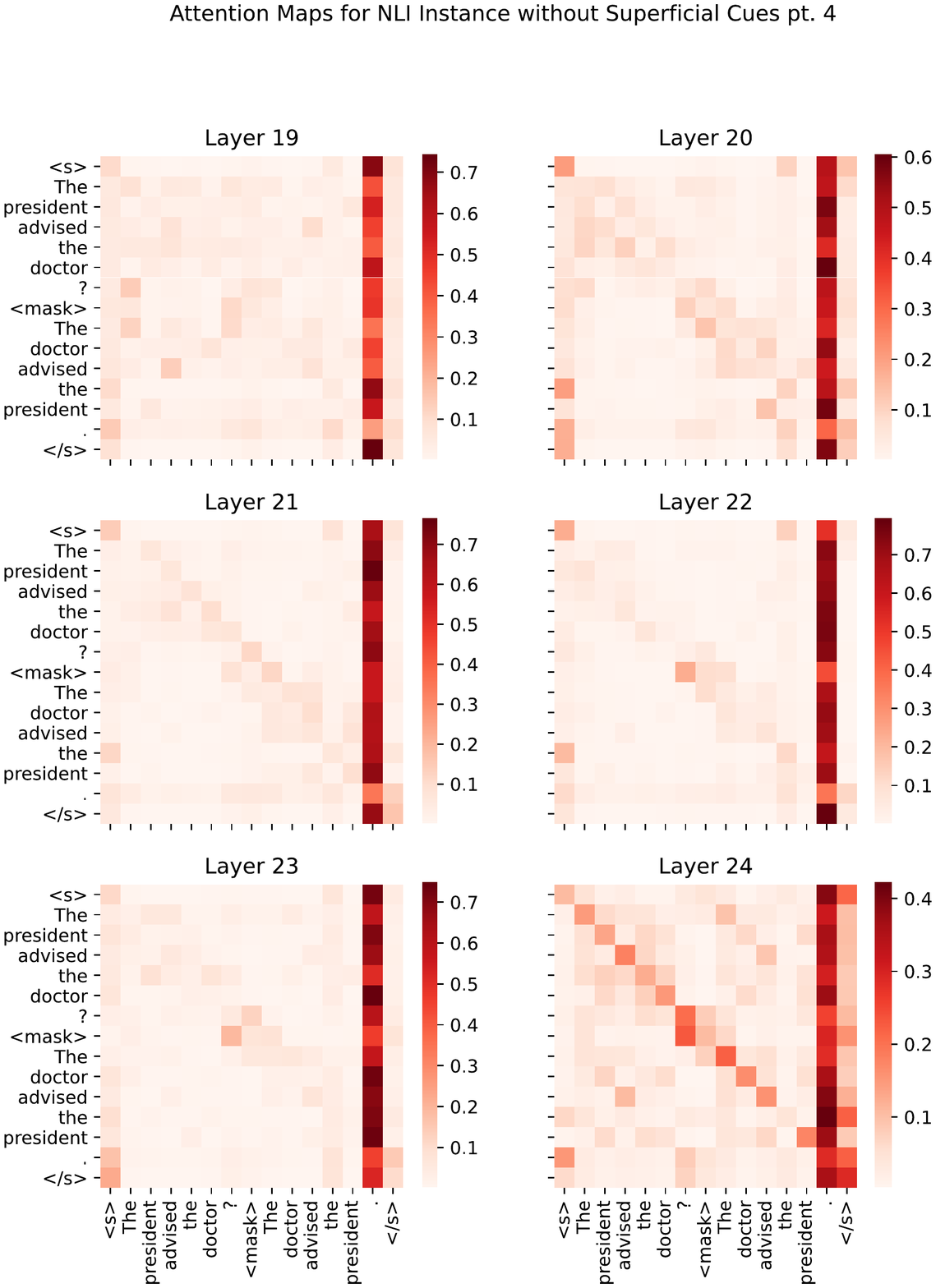}
\end{figure*}
\end{document}